\documentclass[Afour,sageapa,times]{sagej}

\usepackage{moreverb,url}
\usepackage[colorlinks,bookmarksopen,bookmarksnumbered,citecolor=red,urlcolor=red]{hyperref}

\usepackage{amsthm}
\usepackage{amsmath}
\usepackage{amsfonts}
\usepackage{amssymb}
\usepackage{mathrsfs}
\usepackage{bm}
\usepackage{mathtools}
\usepackage{enumerate} 
\usepackage{enumitem}
\usepackage{tensor}
\usepackage{graphicx}
\graphicspath{{figures/}}
\usepackage[table]{xcolor}
\usepackage{multirow}
\usepackage{array, makecell}
\usepackage{balance}
\usepackage{standalone}
\usepackage{tabularx}
\usepackage{ragged2e}
\usepackage{makecell}

\usepackage{color}
\newcommand{\blue}[1]{\textcolor{blue}{#1}}

\theoremstyle{definition}
\theoremstyle{definition}
\newtheorem{remark}{Remark} 
\newtheorem{proposition}{Proposition} 

\usepackage{xparse,soul}
\soulregister\cite7 
\soulregister\citep7 
\soulregister\citet7 
\soulregister\ref7 
\soulregister\pageref7 

\usepackage{xcolor}
\usepackage{lineno}

\renewcommand\hl{} 

\newcommand\BibTeX{{\rmfamily B\kern-.05em \textsc{i\kern-.025em b}\kern-.08em
T\kern-.1667em\lower.7ex\hbox{E}\kern-.125emX}}

\def\volumeyear{2022}

\begin{document}

\runninghead{Zhong et al.}

\title{Integrated Planning and Control of Robotic Surgical Instruments for Tasks Autonomy}

\author{Fangxun Zhong\affilnum{1}, and Yun-Hui Liu\affilnum{1}}

\affiliation{\affilnum{1}T Stone Robotics Institute and Department of Mechanical and Automation Engineering, The Chinese University of Hong Kong, HKSAR, China}

\corrauth{Fangxun Zhong.} \email{fxzhong@cuhk.edu.hk}

\begin{abstract}

Agile maneuvers are essential for robot-enabled complex tasks such as surgical procedures.
Prior explorations on surgery autonomy are limited to feasibility study of completing a single task without systematically addressing generic manipulation safety across different tasks.
We present an integrated planning and control framework for 6-DoF robotic instruments for pipeline automation of surgical tasks.
\hl{We leverage the geometry of a robotic instrument and propose the nodal state space (NSS) to represent the robot state in \textit{SE(3)} space.
Each elementary robot motion could be encoded by regulation of the state parameters via a dynamical system.
This theoretically ensures that every in-process trajectory is globally feasible and stably reached to an admissible target, and the controller is of closed-form without computing 6-DoF inverse kinematics.}
Then, to plan the motion steps reliably, we propose an interactive (instant) goal state of the robot that transforms manipulation planning through desired path constraints into a goal-varying manipulation (GVM) problem.
We detail how GVM could adaptively and smoothly plan the procedure (could proceed or rewind the process as needed) based on on-the-fly situations under dynamic or disturbed environment.
Finally, we extend the above policy to characterize complete pipelines of various surgical tasks.
\hl{Simulations show that our framework could smoothly solve twisted maneuvers while avoiding collisions.
Physical experiments using the da Vinci Research Kit (dVRK) validates the capability of automating individual tasks including tissue debridement, dissection, and wound suturing.
The results confirm good task-level consistency and reliability compared to state-of-the-art automation algorithms.}

\end{abstract}


\maketitle

\section{1. Introduction}

Assisting multi-step dexterous manipulation tasks like minimally invasive surgical procedures is one of the primary directions in robotic applications.
The well-identified advantages of augmented precision and stabilization, compared to manual handling, inherently offer advancement for positioning of surgical instruments.
However, the up-to-date paradigm of robot-assisted minimally-invasive surgery (RAMIS) still requires the surgeon to fully teleoperate the instruments throughout the surgery, or could only automate a single positioning step in ex-vivo environment \shortcite{kwoh1988robot}.
The surgeon's mental concentration remains greatly demanded during the surgery, where the inconsistency in skill proficiency appear among novices and experts.
Once they are automatically planned and executed under surgeon's supervision, the human workload could then be effectively liberated and benefits task effectiveness.

The main goal of surgical task operation is to perform instrument motions and target contact safely according to task guidelines.
\hl{Many surgical tasks in MIS (e.g. tissue dissection and suturing) comprise sequences of highly interactive motion sub-steps towards different targets.}
To cater for the minimally invasive set-up, the instruments usually own a long, slender tool shaft with a wrist-like distal structure to enhance motion dexterity.
\hl{However, task motions commonly encounter highly twisted motions (e.g. consecutive contacts to tissue/needle with inverted orientation).
One or more states in \textit{SO(3)} space might experience large-scale variation(s) ($\sim180^{\circ}$) within a small $R^{3}$ space.
The instruments should also avoid inadvertent intrusion to surrounding tissues and other instruments.}
In RAMIS, these are done by the surgeon using a control interface that transfers his/her delicate hand motions to robot actions.
\hl{To ensure safe interaction to the target (e.g. a firm grasp of a tissue/needle via a specific contact pose), the surgeon also stays aware of the robot trajectory and will only proceed once the tool-target alignment becomes suitable.}
Automating a surgical task with a sequence of such motions demands a smooth and reactive planning control policy to pave the way to reliable task execution.

\hl{The challenges are then to define and ensure motion safety throughout the pipelines across different tasks.
The primary concern is that there must be theoretical proof to support that the robot trajectory is feasible and also reachable to admissible target states.
Each motion should also own re-planning strategy upon specific motion constraints to ensure proper contact to a (moving) target during multi-arm coordination.
Other features including avoidance of obstacle collision and joint limits, real-time implementation, etc., should be solved as well.
More importantly, the above features should be applicable to every elementary task motion throughout the pipeline of surgical tasks to avoid execution-level uncertainties.}
\hl{Planning trajectories for robot manipulators has been widely studied in industrial and/or domestic applications.}
\hl{Prevailing methods include generation of roadmaps} \shortcite{simeon2004manipulation,amato1996randomized}\hl{, sampling-based methods} \shortcite{berenson2009manipulation,berenson2009manipulation} 
\hl{to directly explore the 6-DoF feasible trajectories.}
Meanwhile, on-the-fly planning/control schemes could be more efficient and reactive, e.g. the potential field \shortcite{hwang1992potential} to attract/repel workspace constraints, but is subject to local minima that hinders target reachability.
Aiming for surgery autonomy, most works adopt the above approaches to execute specified motions \shortcite{marinho2019dynamic} and/or a certain task \shortcite{sen2016automating} in surgery.
Learning from demonstration is also utilized to deal with complex maneuvers like knot tying \shortcite{osa2017online} and suturing \shortcite{schulman2013case} to avoid direct path planning.
\hl{We are unaware of any existing works that define generic safety motion constraints emerged from surgical tasks and apply them using a unified framework}.

In this article, we present an integrated planning and control framework for 6-DoF robotic instruments for surgical tasks automation.
The framework covers a globally stable manipulation controller, a reactive manipulation planning policy, and a generic motion primitive to characterize different surgical tasks.
We first classify task-relevant surgical motions into two typical types.
The first type is called \textit{tool-centric actions}, where the robot end-effector is manipulated between (feasible) initial and final configurations without tentative contact to the target but might avoid obstacles.
The second type is the \textit{target-centric actions}, where the end-effector is delicately guided upon certain motion constraints to zero the remaining distance to the target for final contact.
\hl{By leveraging the kinematic property of the instrument's wrist geometry, the nodal vectors could directly parametrize the robot's present-to-goal situation by a new set of parameter space, namely the \textit{nodal state space~(NSS)}.
This avoids generating the (coupled) 6-DoF poses as a high-dimensional nonconvex manifold analysis that hinders globally stable manipulation.
The robot trajectory is then generated and executed upon regulation of the states in NSS encoded by a dynamical system.
The leading robot motion could be rigorously guaranteed stable and smooth under Lyapunov stability theory, and is further globally viable once the state regulation is coordinated by sequential motion allocation (SMA).
We verify its capability to smoothly operate twisted maneuvers, which include drastic orientation changes inevitably occurred in surgical motions.}
Moreover, elementary motions within the task should be coordinated with situation awareness.
To this end, we introduce a dynamic attraction state (DAS), whose dynamics is designed by NSS, as a ``tunable'' instant goal for active guidance of the robot's movements.
This transforms motion autonomy into a \textit{goal-varying manipulation (GVM)} problem, which we could further detail the robot's in-process trajectory through extra motion constraints for path versatility.
We apply GVM as a reactive planning policy to ensure target-centric actions (e.g. repetitive needle grasping in suturing).
\hl{By defining a bounded and robust DAS, GVM could adaptively proceed or rewind the contact motion based on on-the-fly situations.
This could avoid hazardous movements while yielding a constrained path (e.g. unnecessary space intrusion or premature target collision).}
\hl{The framework provides a close-form motion output for tackling online motion constraints with guaranteed trajectory feasibility and target reachability.
Such features are available for each elementary motion without using iterations.
Finally, to elevate task-level generality, we extend such scheme to form different modes of behavior (MoBs), which facilitates depicting and automating pipelines of different tasks.}

The main contributions of this work are summarized as follows:
\begin{itemize}[leftmargin=*]
	\item \hl{A robot state model with skeleton nodes and the NSS for direct formulation of the present-to-goal situation for a robotic instrument based on its kinematic constraints.}
	\item \hl{A DS-based robot controller with synthesized SMA which provides global asymptotic stability proof that guarantees in-process trajectory feasibility and target reachability to an arbitrary (admissible) target.}
	\item \hl{The GVM reactive planning policy to facilitate bilateral transition between target contact and path following to smoothly deal with disturbed targets.}
	\item \hl{The motion primitive and its five MoBs which could define pipelines of different surgical tasks.}
        \item \hl{Simulations and physical experiments on different tasks are conducted for comprehensive validation.}
\end{itemize}

\hl{Our previous works have addressed a few individual problems that target surgery autonomy.
We have proposed an efficient and autonomous robot-camera calibration approach to compute the robotic instrument's pose from a monocular camera for vision-guided instrument manipulation} \shortcite{zhong2020hand}.
\hl{Aiming for autonomous suturing, a monocular-based 6-degree-of-freedom (6-DoF) pose estimation algorithm of a surgical needle has been developed} \shortcite{zhong2016adaptive,zhong2018image}\hl{, which was later implemented to a dual-arm needle insertion control scheme to increase needle insertion accuracy under deformation} \shortcite{zhong2019dual}.

The structure of this article is summarized as follows.
Section 2 reviews the related works from manipulation planning and control approaches to surgery-specific explorations.
Section 3 shows the problem formulation for this work.
Section 4 introduces the new robot kinematics model for the robotic instrument using nodal state space.
Section 5 elaborates the motion-level planning and control strategy and guaranty of safety and in-process constraints.
In Section 6, we elevate the strategy to task-oriented modelling by characterizing different MoBs to form pipelines of different surgical tasks.
Section 7 demonstrates the simulation results concentrating on motion-level performance study, and the experimental results in Section 8 show the overall task-level performance.
Finally, discussions and conclusions are presented in Section 9.

\section{2. Related Works}

\hl{Manipulation planning and control of 6-DoF manipulators subject to in-process constraints has been widely exploited in industrial and domestic applications.
Most works target general workspace constraints, mainly for obstacle avoidance} \shortcite{simeon2004manipulation,stilman2007manipulation} \hl{and/or optimizing characteristics (e.g. avoiding singularities, deadlocks, etc.)} \shortcite{li1997line,ferbach1997method}\hl{, or to decide the sequence of motion profiles for a task (e.g. pushing a box, fetching an object)} \shortcite{billard2019trends,fang2020learning}.
\hl{On the other hand, the motion pipeline required by a surgical task is complicated but clearly given in clinical practice} \shortcite{cao1996task}.
However, the wristed structure of the instrument and limited joint ranges render a smaller size of feasible space manifold and makes it challenging to  coordinate the end-effector and whole-body motions.
Only a few existing works tackle 6-DoF motion control of a robotic instrument, including learning from demonstration to perform a complex but single motion step like looping for knot tying \shortcite{osa2017online}.
\hl{However, it requires collection of totally 4
datasets with $>$ 50 individual demonstrations to cover different conditions for a suture looping motion step.
In }\shortcite{chiu2021bimanual}\hl{, grasping a surgical needle is planned by reinforcement learning, which also depends on 1000 pre-recorded simulated grasps to learn this particular motion step.}
Aiming at task-level autonomy, there are works that introduce convex optimization \shortcite{sen2016automating} and visual servoing \shortcite{pedram2020autonomous} to plan the motions required for needle manipulation in wound suturing.
\hl{However, how to ensure feasible trajectories for twisted maneuvers, which commonly occur when re-orienting a target (e.g. a tissue or needle), are not rigorously provided.}
Similar issues also exist in repetitive target contact like debridement \shortcite{narazaki2006robotic}, where poorly planned motions could easily lead to premature collisions.
As the majority of previous works only focus on a certain step, we will summarize the state-of-the-art approaches for each subproblem in surgical motion autonomy first, and then the works of task-level automation strategies, and their limitations.

\subsection{2.1. Automating Basic Motions}

Pioneer works of robot-enabled instrument manipulation in surgical applications have targeted positioning of an instrument to registered poses, e.g. the tissue cutter in prostatectomy \shortcite{kwoh1988robot} and the biopsy probe in neurosurgery \shortcite{davies1991surgeon}.
Combining clinical CT data, planning an optimal port placement and/or admissible path of the instrument becomes available to provide a safe referenced position for localized treatment \shortcite{schweikard1993motion,adhami2003optimal}.
Among these works, the concept of motion planning is limited to generating a deterministic path using an industrial robot arm for the instrument through admissible space without actual manipulation process or constraints.
Funda et al. \citeyear{funda1996constrained} computed the optimal robot motion of for adjusting the gaze of the laparoscope subject to the remote center-of-motion constraint, which is widely adopted in MIS procedures.
To include online feedback, Wei et al. \citeyear{wei1997real} applied visual servoing to adjust the robot-mounted laparoscope by tracking the instruments from the images.
The method is also adopted in \shortcite{krupa2003autonomous,osa2010framework} to actuate a surgical instrument to a feature-defined target by minimizing image-based errors.
\hl{However, the main disadvantage is that it requires continuous monitoring of visual features during manipulation, which could not robustly define a 6-DoF goal pose during the motion or provide re-planning strategy}.
To deal with multi-instrument environment, Preda et al. \citeyear{preda2015motion} used sampling-based planning scheme (RRT-connected algorithm) for planning collision-free motion step for multiple instruments.
Sozzi et al. \citeyear{sozzi2019dynamic} proposed a DS-based approach with waypoint selection to actuate a non-wristed instrument with collision-free path to another instrument.
The work in \shortcite{marinho2019dynamic} proposed vector fields to guide the motion of a regular instrument with RCM constraint.
Aiming for task-specific motion constraints, a few works adopt the method of learning from demonstration (LfD) to automate complex multi-arm coordinated trajectories like suture looping in knot tying or needle insertion \shortcite{schulman2013case,osa2017online,schwaner2021autonomous} to avoid modelling complex trajectories.
However, it requires pre-recorded motion data for each type of task and is unable to deal with unexpected changes of task process.
Notably, most above works used robot-mounted regular surgical instrument with only 4-DoF motions instead of a wristed robotic instrument.
More recently, the work in \cite{chiu2021bimanual} further introduced reinforcement learning (RL) policy to generate robotic instrument motions.
To feed the online robot states to the system, continuous visual tracking of the instrument features is required via stereo images.

\hl{All previously mentioned works only focused on feasibility study completing a certain motion/task successfully, but failed to define genric safety features to be required by surgery automation to avoid execution-level uncertainties, which is of highest priority in surgeon's perspective} \shortcite{narazaki2006robotic}.
\hl{There also lacks a unified solution to deploy those safety features across different tasks as well.
In this work, we provide a DS-based framework with rigorous proof of global motion stability and trajectory reachability, such that it could handle stable motions to all task-relevant motions.}

\subsection{2.2. Automating Contact Motions}

Interacting surgical instruments to the environment also constitutes a majority of motion steps in surgical tasks.
In MIS, legitimate contact include non-invasive contact (e.g. palpation for tumor localization), grasping (e.g. toward a piece of tissue debris or surgical needle), and invasive contact (e.g. inserting a needle into tissue).
To this end, Patil et al. \citeyear{patil2010toward} addressed manipulation required for tissue retraction by selecting optimal grasping point, but didn't specify the exact motion to safely grasp the tissue in advance.
Since then, there are works addressing 6-DoF instrument-based manipulation of soft tissues for tissue dissection \shortcite{murali2015learning,nagy2018ontology} and deformation control \shortcite{li2020super}.
The works in \shortcite{kehoe2014autonomous,hwang2020applying} automated target grasping in a peg transfer task, which is commonly used in surgical training for novice surgeons.
Among them, the instrument targeting and grasping process were executed separately to avoid unsafe tool-target contact, and the grasping orientations highly resembled each other.
Meanwhile, efforts have been made to needle grasping in wound suturing.
Schulman et al. \citeyear{schulman2013case} introduced LfD to automate manipulation and grasping of a needle with pre-recorded trajectories.
The process had also been automated in terms of direct pick-up \shortcite{d2018automated} and needle hand-off \shortcite{pedram2020autonomous,chiu2021bimanual} using VS and LfD, respectively.
A static needle was targeted and grasped using visual feedback as the motion input, which could reduce reliability when dealing with awkward grasping poses.

\hl{One mutual limitation of these works is that they all regarded grasping as separated steps with a fixed sequence rather than a reactive process, which could be re-planned under sensing/motion disturbances.
The motion constraint during target contact should be embedded as well with the aforementioned safety features, but remains an unmet problem in surgery autonomy.}
We interleave manipulation and contact motion together as a GVM model, such that the robot will adaptively transit between tool-centric or target centric actions with global stability.

\subsection{2.3. Automating Task-Level Procedures}

Completing the pipeline of a surgical task based on step-level automation is the main goal to achieve task-level autonomy.
Researchers have approached various individual surgical tasks with different sub-steps and different controllable DoFs required for the tasks.
Works include autonomous field of view control of endoscope in \shortcite{nageotte2006visual,agustinos2014visual,voros2006automatic,yang2019adaptive} that used VS technique to track a predefined visual feature to allow hands-free adjustment intraoperatively.
In \shortcite{kehoe2014autonomous} and \shortcite{hwang2020applying}, the authors targeted simulated surgical debridement with the instrument trajectories being performed using model predictive control and pose-to-pose interpolations, respectively.
Given the targets' positions, the instrument autonomously manipulated, grasped, and cleared the debris sequentially.
Meanwhile, there are efforts targeting autonomy of tissue-based procedures, including 
non-invasive tasks like tissue palpation \shortcite{nichols2015methods,mckinley2015autonomous} for tumor localization, and invasive tasks like tumor ablation \shortcite{hu2015semi},
tissue dissection \shortcite{murali2015learning,nagy2018ontology}, blood area detection and suctioning from vessel rupture \shortcite{richter2021autonomous}.
Targeting more complex tasks like knot tying, Osa and Van Den Berg \shortcite{osa2010framework,van2010superhuman} automated single knot tying in suturing based on LfD.

Suturing is one of the most common but complicated tasks to be performed in MIS.
It contains a long sequence of motion steps, involving dexterous manipulation of a needle and dual-arm coordination.
Automating such tedious task could significantly reduce human workload, and thus has gained attention from researchers.
For example, the works in \shortcite{kapoor2008constrained,nageotte2005circular}, have planned an optimal trajectory to automate the insertion process of a half-circle surgical needle into the tissue.
A complete stitching automation process was given by \shortcite{nageotte2009stitching} including entry point planning, pose sensing, manipulation, and stitching of the suturing needle.
However, these only constitute incomplete steps required by the suturing guideline.
There are also studies addressing hand-off of the needle between two instruments using human-robot collaborative approach \shortcite{watanabe2016human,watanabe2017single,mikada2020suturing}, and also autonomous approach \shortcite{varier2020collaborative,chiu2021bimanual} using RL-generated motions.
Zhong et al. \shortcite{zhong2019dual} further addressed active tissue deformation during needle insertion to improve insertion accuracy.
Lenard and Shademen \shortcite{leonard2014smart,shademan2016supervised} proposed a suturing device to inherently simplify the suturing task while ensuring a firm stitch.
However, the needle size and stitch width could not be customized restricted by the tool.
Meanwhile, only a few works have completed a throw of wound suturing.
\hl{Sen et al. }\shortcite{sen2016automating} \hl{provided an optimization-based approach to automate a four-throw suturing task, but does not support online trajectory re-planning due to tool-target motion disturbances with theoretical proof for motion stability.}
Pedram et al. \shortcite{pedram2020autonomous} completed an automatic single-throw suturing process using dual-arm coordination using a VS-based control framework.
The instruments are visually tracked during the manipulation, which was reported to own accurate targeting accuracy.
However, continuous visual sensing might hinder its applicability to deal with large orientation change.
\hl{We are unaware of any existing works that provide systematic safety features for complex instrument manipulation in one or more surgical tasks.}

\section{3. Preliminaries}

\subsection{3.1. Problem Formulation}

\hl{Performing a surgical task is essentially to manipulate a robotic instrument from pose to pose in \textit{SE(3)} space across all motion steps.}
Under any (instant) robot configurations, a 6-DoF goal pose of the robot's end-effector (i.e. the instrument's distal tool) is assumed to be available.
It could be either computed from online sensoring feedback or from user input. 
One common method is to add image-based markers on the display (e.g. labeling the tissue's dissecting trajectory or the wound position for suturing), as a surgeon's decision and supervision remains critical to reduce uncertainties of target recognition in supervised robotic surgery \shortcite{haidegger2019autonomy}.
Although computerized surgical image analysis has gained considerable attention to assist target localization \shortcite{loukas2018video}, we emphasize that our aim in this work is an autonomous framework to perform robot motions safely and reliably instead of comprehensive detection algorithms.

\hl{Due to the delicate nature of surgical tasks, we define safety for tasks autonomy to be decomposed into the following two parts: 1) There must be theoretical proof to globally guarantee that the robot motion is admissible in every (future) time instant, and the goal state can be smoothly and stably arrived. The term ``global'' indicates it should apply for any admissible initial/final robot states, and 2) the robot motions must be able to deal with obstacle avoidance (both tool-tool and tool-tissue), joint limit avoidance, etc.
The first part is of primary concern which has not been systematically addressed.
The second part consists of basic safety properties.
We especially target ``twisted'' motions where the distal articulated joints experience large-range adjustment ($>90^{\circ}$), and the robot state especially of \textit{SO(3)} is significantly tuned ($\sim180^{\circ}$), which is normally used in surgery for flipping the grasping orientation of a suturing needle and/or tissue.
Such motion might approach the edge of admissible robot workspace manifold, which is non-convex due to small joint motion range of the two distal joints ($±90^{\circ}$).}
To inherently maximize the performance of system response, the workspace constraints should be embedded into the control model rather than iterative approaches.
The subproblems of instrument non-contact manipulation and contact motion control are processed under the same core in an integrated DS, and must be applicable to all motions appeared in one or more tasks.

\subsection{3.2. Nomenclature}

We regard the entire structure of the detachable tube-like instrument and its proximal motorized joints (totally 7 DoFs) as a surgical robot (or just ``robot'' if not further clarified).
We call the distal structure of the instrument to achieve clamping, grasping, etc. as the ``tool'', which is also the end-effector in robotics.
\hl{To detail the workspace control problem, we use the robot's 6-DoF end-effector pose (denoted by a 3D Cartesian coordinate frame $\mathcal{F}$) or its corresponding state $\mathcal{X}$ (computed from a newly proposed set of space parameters) to represent the robot's present or goal configuration.
The acronyms appeared in this article have been listed out in Table \ref{T1}.}

\begin{table}[t]
\small\sf
\caption{\hl{List of acronyms used in this article (in alphabetical order)}.} \label{T1}
\begin{tabularx}{0.47\textwidth}{r|l}
\toprule
~~~~~~~~~~~Acronym & Definition~~~~~~~~\\
\midrule
\vspace{0cm}
DAS & Dynamic Attraction State\\
\vspace{0cm}
DHc & Denavit-Hartenberg convention\\
\vspace{0cm}
DoF & Degree-of-freedom\\
\vspace{0cm}
DS & Dynamical system\\
\vspace{0cm}
GVM & Goal-varying manipulation\\
\vspace{0cm}
LfD & Learning from demonstration\\
\vspace{0cm}
MIS & Minimally invasive surgery\\
\vspace{0cm}
MoB & Mode of Behavior\\
\vspace{0cm}
NSS & Nodal state space\\
\vspace{0cm}
RAMIS & Robot-assisted minimally invasive surgery\\
\vspace{0cm}
RCM & Remote center-of-motion\\
\vspace{0cm}
RL & Reinforcement learning\\
\vspace{0cm}
SMA & Sequential motion allocation\\
\vspace{0cm}
SMP & Surgical motion primitive\\
\vspace{0cm}
VS & Visual servoing\\

\bottomrule
\end{tabularx}\\[0pt]

\end{table}

\begin{figure}[t]
\centering
\includegraphics[width=0.48\textwidth,keepaspectratio]{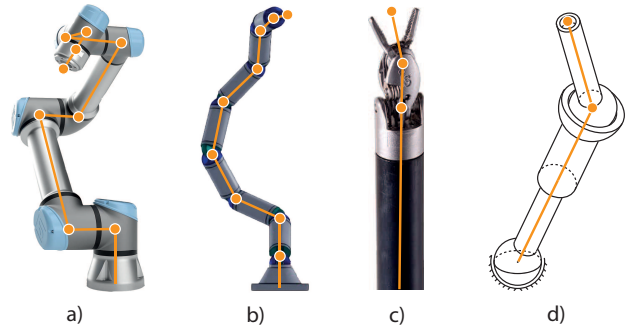}
\caption{The definition of ``nodes'' on different types of serial robot manipulators. a) the UR5 as a typical robot arm; b) a hyper-redundant robot arm \protect\shortcite{li2015ultimate}; c) the EndoWrist as a representative of robotic surgical instrument; d) a highly wristed non-redundant robot with only three links (with DoFs being 3R-P-2R).}
\label{fig:Nodes}
\end{figure}

\section{4. Robotic Instrument Model}

We begin by derive the kinematics of a robotic surgical instrument based on its mechanical properties.
A typical serial-link manipulator generally owns six joints to fully determine the 6-DoF pose of the end-effector (\textit{SE(3)} space) in 3D Cartesian space modelled by a coordinate frame $\mathcal{F}_{e}(\mathcal{O}_{e},x_{e},y_{e},z_{e})$, which could also describe the robot's state space $\textbf{x}\in\mathbb{R}^{6}$ in robot kinematics with the following well known form:
\begin{equation} \label{eq:x}
	\textbf{x} = \textbf{f}(\textbf{q})
\end{equation}
with $\textbf{q}=~[q_{1}~\ldots~q_{6}~]^{\intercal}$ denotes the generalized coordinate vector in joint space that describes the robot joint angles.
As robot joints are mostly revolute, the mapping $\textbf{f}(\cdot)$ is usually nonlinear and indicates a coupled relationship between the end-effector's position and orientation while being regulated.
Such property remains in velocity mapping using the Jacobian matrix $\textbf{J}(\textbf{q})\in\mathbb{R}^{6\times6}$:
\begin{equation} \label{eq:v}
    \dot{\textbf{x}} = \textbf{J}(\textbf{q})\dot{\textbf{q}}
    .
\end{equation}
Then, adjustment of the end-effector's pose will inevitably lead to movements for all six joints.
Considering a complex motion step, the more ``twisted'' trajectory the end-effector undergoes, the more details are to be planned for describing its profile.
Moreoever, the limited joint motion ranges ($\pm90^{\circ}$ for the last two joints) results in a nonconvex reachable workspace manifold, where the path should be further refined to avoid all infeasible regions.
Existing approaches deal with such trajectory complexity using \textit{a priori} workspace analysis \shortcite{mirrazavi2018unified} or iterative optimization \shortcite{zucker2013chomp}, which could not guarantee real-time and/or theoretically feasible output in such high-dimensional space, and are unsuitable for surgical applications where the environment is usually temporarily set up and must not sacrifice efficiency.
Thus, a new parametrization method is to be developed to reformulate the kinematic relationship between the robot's proximal parts and its end-effector. 
Here, we introduce the concept of \textit{skeleton node analysis} that regards each junction (or the ``elbow") between two consecutive links (or the ``limbs") along the robot skeleton as a point of interest.
Specifically, we define a 3D point $p\in\mathbb{R}^{3}$ in Cartesian space as a skeleton node of the robot if it satisfies either of the following criteria:
\begin{itemize}[leftmargin=*]
    \item It is the intersection point of the rotary axes of two consecutive robot joints.
    \item It is the intersection point of the rotary axis of a robot joint and the geometric centerline of a link (regarding straight rigid links).
    \item It is the origin of the end-effector's 3D Cartesian coordinate frame.
\end{itemize}
Meanwhile, we disqualify a candidate if it overlaps an existing node which has already been defined from proximal robot parts.
The general form of a node set considering a 6-DoF serial robot manipulator could be given as follows:
\begin{equation} \label{eq:N}
	\mathcal{N} = 
	\begin{Bmatrix}
		n_{1} & n_{2} & \ldots & n_{k}
	\end{Bmatrix}
\end{equation}
where the dimension of the node set $\mathcal{N}$ being $k\in[2,~12]:k\in\mathbb{Z}^{+}$, which is associated with the number of non-zero link length/offsets due to different kinematic design and does not necessarily correlates to the number of joints (see Fig. \ref{fig:Nodes} for the examples of $\mathcal{N}$ on different types of robots). 
This implies the skeleton node analysis emphasizes an explicit 3D geometry of the robot's kinematic chain relative to the end-effector, which is not achievable by using the Denavit$-$Hartenberg convention (DHc).
Meanwhile, one can denote $n$ 3D position unit vectors as a set $\mathcal{V} =\{\nu_{1}~\nu_{2}~\ldots~\nu_{k}\}$ referenced from the corresponding $n$ skeleton nodes and head to arbitrary directions.
Each vector, named as the \textit{nodal vector}, is described in robot base frame and is virtually fixed to the skeleton and thus is adjustable by all joints proximal to the attached node.
Such group of nodal vectors are then denoted by $\Xi\in\mathbb{R}^{k\times3}$, with
\begin{equation} \label{eq:xi}
	\Xi =
	\begin{bmatrix}
		\boldsymbol{\xi}_{1} & \boldsymbol{\xi}_{2} & \ldots & \boldsymbol{\xi}_{k}
	\end{bmatrix}
	.
\end{equation}
%
\begin{figure}[t]
\centering
\includegraphics[width=0.45\textwidth,keepaspectratio]{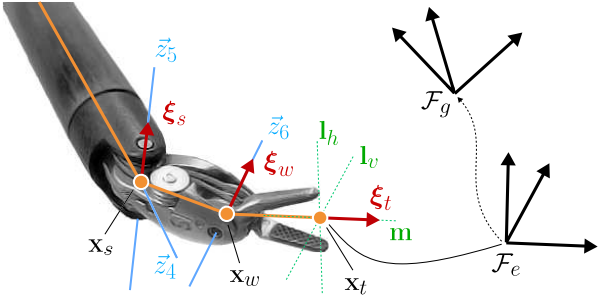}
\caption{Kinematics of the wristed instrument and the illustration of the nodes and nodal vectors according to our definition.}
\label{fig:EndoWrist}
\end{figure}
Normally, for an arbitrary node $n_{i},~i\in[1,k]$, there always exists a nonlinear and variable transformation between it and the end-effector (except $\nu_{k}$) regarding its distal joints.
No consistent geometric relationship between the nodal vectors and the end-effector could be derived.

We then use the skeleton nodes to analyze a robotic surgical instrument implemented to MIS.
Without loss of generality, we consider the structure of the EndoWrist (by Intuitive Surgical Inc) which has been widely adopted for designing robotic surgical systems, and regard it as a typical example to study surgery autonomy\footnote{Examples of robotized surgical instruments that adopt the design of the wristed structure include da Vinci Xi (Intuitive Surgical), Versius (CMR Surgical), Micro Hand S (Weigao Group Medical), RAVEN (Applied Dexterity), etc.}.
The instrument owns a slender cylindrical shaft that provides 4-DoF RCM-constrained motions to yield the minimally invasive set-up, a 2-DoF wristed joint set to provide wristed motions, and one extra DoF for tool actuation.
Such kinematics leads its skeleton node set to be degenerated into the following form sufficed by $k=3$:
\begin{subequations} \label{eq:N-ins}
\begin{align}
	\mathcal{N} = &
	\begin{Bmatrix}
		n_{1} & n_{2} & n_{3}
	\end{Bmatrix}
	\\
	\boldsymbol{\Xi} = &
	\begin{Bmatrix}
		\boldsymbol{\xi}_{1} & \boldsymbol{\xi}_{2} & \boldsymbol{\xi}_{3}
	\end{Bmatrix}
	\end{align}
\end{subequations}
where $\boldsymbol{\Xi}_{ins}\in\mathbb{R}^{3\times3}$, and the nodes $n_{1}$, $n_{2}$, $n_{3}$ geometrically represent the shaft-to-wrist junction $\textbf{x}_{s}\in\mathbb{R}^{3}$, the wrist-to-tool junction $\textbf{x}_{w}\in\mathbb{R}^{3}$, and the tool tip $\textbf{x}_{t}\in\mathbb{R}^{3}$, respectively, as shown in Fig. \ref{fig:EndoWrist}.
Then, due to the 3D geometric constraint of a wristed robotic instrument, one can define the following orthogonality that unconditionally holds:

\begin{equation} \label{eq:per}
	\begin{bmatrix}
		\boldsymbol{\xi}_{1}^{\intercal} & \ldots & \textbf{0}_{1\times3} \\
		 \vdots & \boldsymbol{\xi}_{2}^{\intercal} & \vdots \\
		 \textbf{0}_{1\times3} & \ldots & \boldsymbol{\xi}_{3}^{\intercal}
	\end{bmatrix}
	\begin{bmatrix}
		\textbf{l}_{h} \\
		\textbf{l}_{v} \\
		\textbf{m}
	\end{bmatrix}
	= \textbf{0}_{9\times1}
\end{equation}
where $\textbf{m}$ indicates the longitudinal unit vector (or heading) and $\textbf{l}_{h}$, $\textbf{l}_{v}$ being two lateral unit vectors with respect to the tool's heading orientation (refer to Fig. \ref{fig:EndoWrist} for illustration).
If one considers a typical end-effector frame $\mathcal{F}_{e}(\mathcal{O}_{e},x_{e},y_{e},z_{e})$ using DHc whose origin position overlaps $\textbf{x}_{t}$, the following factorization could also be derived:
\begin{equation} \label{eq:H}
	\begin{bmatrix}
		\textbf{l}_{h} & \textbf{l}_{v} & \textbf{m}
	\end{bmatrix}
	=
	\textbf{H}
	\begin{bmatrix}
		\vec{x}_{e} & \vec{y}_{e} & \vec{z}_{e}
	\end{bmatrix}
\end{equation}
where $\textbf{H}\in\mathbb{R}^{3\times3}$ is the constant orthogonal matrix computed by a series of fundamental transformations.
Notably, this reveals for any wristed robot manipulators, its end-effector orientation could be interpreted by intermediate body parts (characterized by nodal vectors) via a constant geometric relationship.\footnote{While satisfying (\ref{eq:H}), for sake of simplicity, we define the node vectors and the end-effector frame axes (using DHc) such that $\textbf{H} =\textbf{I}_{3\times3}$ and $k_{s}=k_{w}=k_{t}=1$.}
The nodal vectors could also be depicted as $\boldsymbol{\xi}_{1}=k_{1}\vec{x}_{e},~\boldsymbol{\xi}_{2}=k_{2}\vec{z}_{e},~\boldsymbol{\xi}_{3}=k_{3}\vec{y}_{e}$. 
This is an important kinematic property for us to study the robot's whole-body motion coordination subject to task-relevant constraints.
Here, to yield (\ref{eq:per}), we could select one group of nodal vectors for the instrument:
\begin{equation} \label{eq:axis}
    	\boldsymbol{\xi}_{1} = \vec{z}_{6}~~~~~~
    	\boldsymbol{\xi}_{2} = \vec{z}_{5}~~~~~~
    	\boldsymbol{\xi}_{3} = \vec{x}_{6}
\end{equation}
where $\vec{z}_{5},\vec{z}_{6}$ denote the axial direction of the respective joint $q_{5}$ and $q_{6}$ from robot base, and $\vec{x}_{6}$ being along the tool's heading direction.

We decompose the robot joints correspondingly into the following:
\begin{equation} \label{eq:q}
    \textbf{q} = 
	\begin{bmatrix}
		\textbf{q}_{s}
		& q_{4}
		& q_{5}
		& q_{6}
	\end{bmatrix}
	^{\intercal}
\end{equation}
where $\textbf{q}_{s}=[q_{1},q_{2},q_{3}]\in\mathbb{R}^{3}$ denotes the joints that control the shaft pose, and the rest regulate the wristed part of the instrument.
Then, combining DHc to the skeleton nodes, we further give the derivation of the nodal vectors computed with respect to the robot base frame:
\begin{equation} \label{eq:xi}
    \boldsymbol{\xi}_{i} = \textbf{R}_{s}(\textbf{q}_{s})\prod_{i = 1}^{3} {}^{i+2}\textbf{R}_{i+3}(q_{i+3})\overline{\boldsymbol{\xi}}_{i},~~~i=1,2,3
\end{equation}
where $\textbf{R}(\cdot)\in\mathbb{R}^{3\times3}$ denotes the rotation matrices and $\overline{\boldsymbol{\xi}}_{1,2,3}$ are the nodal vectors described in local frames.
In the next section, we will implement $\boldsymbol{\xi}_{i}$ as a new media for the robot's present-to-goal model and for versatile motion control of the robot's end-effector as well as the body skeleton.


\section{5. Motion-Level Autonomy Architecture}

Based on the above modelling, we first aim to regulate the robot's end-effector to reach an arbitrary (but feasible) goal pose by proposing a basic control architecture to provide safety features to the robot's motions.
We then extend it to yield more task-relevant motion constraints and utilize the derived properties to integrate a series of dexterous maneuvers into a single motion step.

\subsection{5.1. DS-Guided Instrument Manipulation}
\subsubsection{5.1.1. End-Effector Control System \\}
The DS is an effective mathematical approach to generate stable, convergent robot motions which are efficiently reactive to instant state variations and/or disturbances.
\hl{This endows DS-guided manipulation with powerful adaptability in motion re-planning without the need to analyze the entire trajectory (compared to optimization-based approaches), which is superior to tackle robot manipulation once the motionguconstraints are embedded into the system model.}

\begin{figure}[t]
\centering
\includegraphics[width=0.48\textwidth,keepaspectratio]{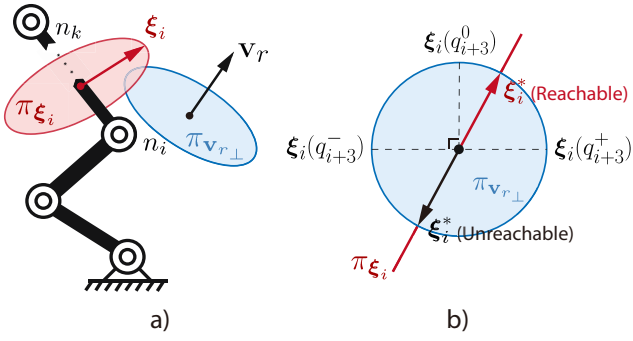}
\caption{Geometric representation of the variability of $\boldsymbol{\Xi}$ subject to $q_{i+3}$ in a), and the solution uniqueness to achieve $\textbf{v}_{r}{}^{\intercal}\boldsymbol{\xi}_{i}(q_{i+3})=0$ when considering the physical joint motion of $q_{i+3}$ in 3D workspace in b).
Respectively, $q_{i+3}^{-}$, $q_{i+3}^{0}$ and $q_{i+3}^{+}$ denotes the negative limit, zero position, and positive limit.}
\label{fig:UniqueAttractor}
\end{figure}

\hl{We aim to incorporate the instrument's kinematic property using NSS analysis to the DS to enable globally-guaranteed safe instrument manipulation planning.}
We start by giving the general form of a time-invariant nonlinear dynamical system as follows:
\begin{equation} \label{eq:DS-basic}
	\dot{\boldsymbol{\zeta}}(t) = \textbf{g}(\boldsymbol{\zeta}(t),\textbf{u}(t))
\end{equation}
which indicates the evolution of the system's states $\boldsymbol{\zeta}(t)$ subject to the control input $\textbf{u}$, and $\textbf{g}(\cdot)$ denotes the differentiable vector-valued function.
To apply such technique to robot's motion generation, the robot states $\boldsymbol{\zeta}(t)$ must be appropriately defined to enable its dynamics (via $\textbf{g}(\cdot)$) to guide the robot with demanded performances as the system evolves over time.
The states are usually defined as the 6-DoF end-effector pose or configuration space \shortcite{mirrazavi2016dynamical} and thus $\textbf{g}(\cdot):\mathbb{R}^{6l}\mapsto\mathbb{R}^{6}$.
Here, unlike the above methods, we use the nodal vectors to encode the relationship between the robot's present and goal configurations based on its body parts.
Consider an arbitrarily defined $\mathcal{F}_{g}(\mathcal{O}_{g},x_{g},y_{g},z_{g})$ as the target pose to reach, we particularly introduce a new parametrization space $\mathcal{H}$ which is defined as follows:
%
\begin{equation} \label{eq:eta}
	\boldsymbol{\eta}(\textbf{q}) =
	\begin{bmatrix}
	    \eta_{1} \\ \eta_{2} \\ \eta_{3} \\ \eta_{4}
	\end{bmatrix}
	:=
	\begin{bmatrix}
		 d^{*}(\textbf{x}_{s},~\textbf{x}_{s_{g}}) \\ 
		 \langle\boldsymbol{\xi}_{1},~\vec{x}_{g}\rangle \\
		 \langle\boldsymbol{\xi}_{2},~\vec{z}_{g}\rangle \\
		 \langle\boldsymbol{\xi}_{3},~\vec{y}_{g}\rangle
	\end{bmatrix}
	\subseteq\mathcal{H}\subseteq\mathbb{R}^{4}
\end{equation}
where the parameters $\eta_{2},\eta_{3},\eta_{4}$ depict via inner product the orientation alignment of the end-effector from its target, $\textbf{x}_{s_{g}}$ is the goal position of $\textbf{x}_{s}$ computed from $\mathcal{F}_{g}$.
The scalar $d^{*}(\cdot)=\textbf{x}_{s}^{\intercal}(\textbf{x}_{s} - \textbf{x}_{s_{g}})/\vert\vert\textbf{x}_{s} - \textbf{x}_{s_{g}}\vert\vert^{2}$ defines the signed distance of $\textbf{x}_{s}$ from $\textbf{x}_{s_{g}}$ via vector projection.
Note that, getting $\textbf{x}_{s_{g}}$ requires computation of inverse kinematics which could not be analytically solved due to the coupled position and orientation.
Here, we compute a ``naive" but also closed-form $\tilde{\textbf{x}}_{s_{g}}$
\begin{equation} \label{eq:x_sg}
	\tilde{\textbf{x}}_{s_{g}} = \textbf{x}_{g}-\textbf{x}_{t}+\textbf{x}_{s}
\end{equation}
\hl{from the knowledge of distal joints (encoded by $\textbf{x}_{t}(\textbf{q}$) which are uncontrollable by $\textbf{x}_{s}(\textbf{q}_{s})$.
Note that in (\ref{eq:x_sg}), $\tilde{\textbf{x}}_{s_{g}}$ is derived based on the instant forward kinematics, which provides ``coarse'' guidance towards the end-effector's goal position using only $\mathbf{q}_{s}$.
However, we will later show that such $\tilde{\textbf{x}}_{s_{g}}$ does not affect the performance of the target DS and could converge to its genuine value subject to robot control.}
We name $\mathcal{H}$ as the \textit{nodal state space (NSS)} which characterizes the robot's goal-oriented spatial relationship using the skeleton features of the nodes.
The robot's current state $\mathcal{X}$ and goal state $\mathcal{X}_{g}$ could now be reformulated by the components in (\ref{eq:eta}) as follows
\begin{equation} \label{eq:state}
    \begin{aligned}
        \mathcal{X} & := 
        \begin{bmatrix}
        \textbf{x}_{s} & \boldsymbol{\xi}_{1} & \boldsymbol{\xi}_{2} & \boldsymbol{\xi}_{3}
        \end{bmatrix}
    	^{\intercal}
        , \\
    	\mathcal{X}_{g} & := 
        \begin{bmatrix}
            \textbf{x}_{s_{g}} & \vec{x}_{g} & \vec{y}_{g} & \vec{z}_{g}
        \end{bmatrix}
    	^{\intercal}
    	.
    \end{aligned}
\end{equation}
Now, we will use ``state'' in (\ref{eq:state}) to describe the robot's status which is equivalent to a 3D Cartesian space, as we aim to solve automation via Cartesian workspace analysis.
Several properties that NSS could contribute to robot motion guidance are highlighted.
\proposition{$\boldsymbol{\eta}(\textbf{q})$ has a globally unique equilibrium $\boldsymbol{\eta}^{*}$ with $\boldsymbol{\eta}^{*}=\textbf{0}$.\\
\textit{Proof}: Regarding $\eta_{1}$, any given (feasible) $\textbf{x}_{s_{g}}$ will correspond to a globally unique solution of $\textbf{q}_{s_{g}}$, as the local kinematics $\textbf{x}_{s}=\textbf{f}_{s}(\textbf{q}_{s})$ Cartesian workspace manifold being a solid sphere \shortcite{from2013kinematics}.
\hl{This shows that $\textbf{f}_{s}(\cdot)$ is convex and non-singular for regulating $\eta_{1}$.
Meanwhile, according to (\ref{eq:eta}), there exists multiple solutions of $\boldsymbol{\xi}_{i+1}$ to satisfy $\eta_{i}=0$ ($i=2,3,4$).
However, we will prove in Proposition 2 that due to the joint limits of the instrument, only one $\boldsymbol{\xi}_{i+1}$ is physically reachable subject to $\textbf{x}_{s}\rightarrow\textbf{x}_{s_{d}}$, which further corresponds to a globally unique $q_{i+3}$.}
Thus, there exists a unique $q$ to meet $\eta_{i}=0$ for $i=1,2,3$ for any feasible $\mathcal{X}_{g}$ interpreted by $\mathcal{F}_{g}$.}

\proposition{Given a 3D unit vector $\textbf{v}_{r}$ based on $\mathcal{F}_{g}$ and a nodal vector ${\boldsymbol{\xi}_{i}}(\textbf{q}_{s},q_{i+3})$ of node $n_{i}$ ($i=1,2,3$), there exists a unique solution $q_{i+3}^{*}$ for $\boldsymbol{\xi}_{i}$ such that $\textbf{v}_{r}{}^{\intercal}\boldsymbol{\xi}_{i}(\textbf{q}_{i+3})=0$ subject to $\textbf{q}_{s}\rightarrow\textbf{0}$.\\
\noindent{\textit{Proof}:} 
According to (\ref{eq:xi}), the derivation of the nodal vector $\boldsymbol{\xi}_{i}$ is simplified to the following:
\begin{equation} \label{eq:prop2_1}
    \boldsymbol{\xi}_{i}=\textbf{C}{}^{i+2}\textbf{R}_{i+3}(q_{i+3})\overline{\boldsymbol{\xi}_{i}}~~~s.t.~~~\textbf{q}_{s}\rightarrow\textbf{0}
\end{equation}
\hl{where the variation of the rotation matrix $\textbf{C}\in\mathbb{R}^{3\times3}$ is negligible due to the settled proximal robot parts.
$\textbf{R}_{i+3}$ is the fundamental z-axis rotation matrix which acts as a periodical factor applied to $\overline{\boldsymbol{\xi}_{i}}$ and is solely regulated by $q_{i+3}$.
Thus, $\boldsymbol{\xi}_{i}$ has a period of $2\pi$ subject to change of $[q_{i+3},~ q_{i+3}+2\pi]$, as its components own the following form:}
\begin{equation} \label{eq:prop2_2}
	\xi_{i,j} = a_{1} + a_{2}\text{cos}(q_{i+3}) + a_{3}\text{sin}(q_{i+3}),~~~~j=1,2,3
\end{equation}
\hl{where $a_{1},a_{2},a_{3}$ are constant scalars. Therefore, there are two possible solutions of $q_{i+3}$ for $\boldsymbol{\xi}_{i}=\textbf{v}_{r}{}^{\intercal}\boldsymbol{\xi}_{i}(\textbf{q}_{i+3})=0$ within $2\pi$.
For surgical robots, the motion ranges of the two distal joints $\textbf{q}_{w}$ and $q_{t}$ are limited to $[-\pi/2,~\pi/2]$.
Thus, it is impossible to generate two toggled directions of $\boldsymbol{\xi}_{i}$.
There exists only one $q_{i+3}$ that satisfies $\textbf{v}_{r}{}^{\intercal}\boldsymbol{\xi}_{i}=0$ within the robot's feasible workspace (i.e. $[q_{i+3}^{-},q_{i+3}^{+}]$) to yield $\eta_{i}=0$ (shown in Fig. \ref{fig:UniqueAttractor}b).
That corresponds to the unique solution of $\boldsymbol{\xi}_{i}$, denoted by $\boldsymbol{\xi}_{i}^{*}$.
Meanwhile, the direction of vector $\boldsymbol{\xi}_{i}$ periodically changes through $[q_{i+3},~ q_{i+3}+2\pi]$, which sweeps $\boldsymbol{\xi}_{i}$ through a circle whose center being exactly the position of $n_{i}$ within a plane $\pi_{\boldsymbol{\xi}_{i}}$ (see Fig. \ref{fig:UniqueAttractor} for geometrical interpretation).}} \footnote{The approximation is ensured by an additional assumption that $\textbf{x}_{s}$ 
must be continuously and stably guided to $\textbf{x}_{s_{g}}$.
We will shortly provide a DS that regulates $\textbf{q}_{s}$ to satisfy such performance.}

Now, we establish a dynamical system to guide the instrument end-effector to its target pose.
Define the NSS $\mathcal{H}\subseteq\mathbb{R}^{4}$ as the state variables, then due to Proposition 1, there exists a globally unique equilibrium at origin, i.e. $\boldsymbol{\eta}^{*}=\textbf{0}_{4\times1}$.
$\textbf{u}$ is of $\dot{\textbf{q}}$ or $\textbf{q}$ only.
Thus, we propose the following controller:
\begin{equation} \label{eq:ds-u}
    \textbf{u}=-\left(\frac{\partial\boldsymbol{\eta}(\textbf{q})}{\partial\textbf{q}}\right){}^{-1}\boldsymbol{\Gamma}\boldsymbol{\eta}(\textbf{q})
\end{equation}
where $\textbf{M}$ is the $\boldsymbol{\Gamma}\in\mathbb{R}^{6\times6}\succ0$ is the diagonal gain matrix applied to joint velocities.
Then, by applying the control input $\textbf{u}=\dot{\textbf{q}}$ and substitute (\ref{eq:ds-u}) to $\boldsymbol{\dot{\eta}}(\textbf{q})$ leads to the following:
\begin{equation} \label{eq:ds-ds}
    \boldsymbol{\dot{\eta}}=-\boldsymbol{\Gamma}\boldsymbol{\eta}
\end{equation}
which clearly indicates an asymptotically stable DS (with sole negative real parts in phase plane analysis) whose the attractor is exactly $\boldsymbol{\eta}^{*}=\textbf{0}_{4\times1}$.
Its evolution guides the instrument motions via $\dot{\textbf{q}}(t)$ to reach the end-effector $\mathcal{F}_{e}$ to any feasible $\mathcal{F}_{g}$.
As $\boldsymbol{\eta}$ is also bounded due to (\ref{eq:eta}), from (\ref{eq:ds-ds}), the DS that regulates $\boldsymbol{\eta}$ is globally asymptotically stable from Lyapunov stability theory as each equilibrium in $\boldsymbol{\eta}^{*}$ is globally unique (refer to Proposition 2) \shortcite{slotine1991applied}.
The system is also of $C^{1}$ continuity as (\ref{eq:x}) and (\ref{eq:xi}) are all differentiable regarding both $\dot{\textbf{q}}(t)$ and $\textbf{q}(t)$.

\remark{Deploying the controller (\ref{eq:ds-u}) to robot's target reaching control require interpretation of a $\mathcal{F}_{g}$ to $\boldsymbol{\eta}$.
$\mathcal{F}_{g}$ is normally provided as a 3D Cartesian coordinate frame for task-oriented robot manipulation.
Then, according to (\ref{eq:per}) and (\ref{eq:H}), $\boldsymbol{\eta}_{2,3,4}$ are all computable.
Meanwhile, we use $\tilde{\textbf{x}}_{s_{g}}$ as in (\ref{eq:x_sg}) as approximation of $\textbf{x}_{s_{g}}$ to avoid computing inverse kinematics, which is not differentiable for on-the-fly manipulation guidance.
}

\subsubsection{5.1.2. Sequential Motion Allocation (SMA) \\}

The current form of DS only addresses the convergent motion of the end-effector to a goal pose and does not necessarily guarantee the feasibility of the in-process trajectory.
Surgical tasks commonly require grasping the needle and tissue from specific poses to minimize potential trauma.
When transitioning between awkward poses, the robot might encounter joint limits and/or near-singular configurations.
In terms of a serial robot manipulator, one robot joint could parallelly tune different $\eta$.
Therefore regulating a specific $\eta_{i}$ might cause disturbanceS to other (distal) $\eta_{j},~j>i,~j\in\mathbb{N}^{+}$, which could decrease the overall control effectiveness and target reachability.

To solve the above issues, new constraints must be added to the system improve the instrument's manipulation delicacy.
Here, we introduce the \textit{sequential motion allocation (SMA)} that progressively allocates the regulation output from proximal to distal robot joints.
This will facilitate ``rhythmic" movements of the robot's body parts and avoid "clumsy" joint motions, which could be harmful to both the robot and the environment.
\hl{We achieve SMA by proposing a new set of state-based parameters $\lambda\in\mathbb{R}$ as the nodal allocation coefficients} to dynamically weight the control output deployed to $\eta$ for each node based on (\ref{eq:eta}) and (\ref{eq:ds-u}).
The definition of $\lambda$ is chosen by considering the following properties to meet the task-relevant motion characteristics:
\begin{enumerate}[leftmargin=*]
    \item {We set all coefficients to $\lambda\in[0,1]\subseteq\mathbb{R}$, $\lambda\in\mathcal{C}^{0}$ to normalize the scales of all $\eta$ among different between types of robot joints and/or their different ranges of motion during regulation.
    $\lambda=1$ and $\lambda=0$ indicates the motion output that regulates the corresponding $\eta$ has been fully deployed and suspended, respectively.
    }
    \item {To achieve sequential actuation of the skeleton nodes, the $\lambda_{i}$ that allocates the regulating motions to the current $\eta_{i}$ should initiate posterior to the elevation of the previous (proximal) $\lambda_{i-1}$.
    This is interpreted as $\lambda_{i-1}\geq\lambda_{i}~~\forall \lambda_{i-1},\lambda_{i}\in[0,1]$}.
    \item {To incorporate SMA to whole-body coordination, $\lambda_{i}$ needs to open a gap to the rising $\lambda_{i-1}$ ahead, but also needs to catch up with $\lambda_{i-1}$ as $\lambda_{i-1}\rightarrow1$ to eventually enable full regulation.
    This indicates the boundary conditions of $\lambda_{i}(x)$ to yield $\dot{\lambda}_{i}(x)|_{x=0}=0~\forall i$ and $\dot{\lambda}_{i}(x)|_{x=1}=1,~\forall i>1$ with $x=\lambda_{i-1}$}.
\end{enumerate}
All $\lambda$ will constitute a space $\mathcal{L}$ whose dimension corresponds is equal to that of $\mathcal{H}$, with $\lambda=[\lambda_{1},\lambda_{2},\lambda_{3},\lambda_{4}]^{\intercal}\in\mathcal{L}\subseteq\mathbb{R}^4$.

A feasible mathematical form of $\lambda$ to achieve SMA could be given as follows:
\begin{equation} \label{eq:lambda}
    \lambda_{i} = 
    \begin{cases}
        \exp(-k_{1}\vert\vert\eta_{1}\vert\vert), & i=2 \\
        \exp(-k_{i}(1/\lambda_{i-1}-1)), & i > 2
    \end{cases}
\end{equation}
where $k_{i}$ is the preset gain for each $\lambda_{i}$.
We set $k_{i}=1~\forall k>1$ to meet $\lim_{\lambda_{i-1}\rightarrow0}\dot{\lambda}_{i}(\lambda_{i-1})=0~\forall i$.
The definitions in (\ref{eq:lambda}) shows that, to initiate SMA, $\lambda_{1}$ needs to be exclusively modelled apart from the others.
Once $\eta_{1}\rightarrow0$, the control output will be smoothly allocated to the next node with $\eta_{2}$ by elevating $\lambda_{2}$, and vice versa.
The subsequent $\lambda_{i}$ are tuned directly based on $\lambda_{i-1}$ to ensure a robust and flexible SMA process regardless of the robot's initial/final state or its control input.
\footnote{Such setting of $\lambda$ could guarantee a sequenced actuation process using SMA especially when the instrument experiences weak manipulability and/or extreme joint positions which could result in drastic joint control input.}
\begin{equation} \label{eq:ds-u-bar}
    \overline{\textbf{u}}=\textbf{L}\textbf{u}
\end{equation}
with
\begin{equation} \label{eq:L}
\mathbf{L}
=
	\begin{bmatrix}
		1 & \mathbf{0}_{1\times3} \\
		\mathbf{0}_{3\times1} & diag(\boldsymbol{\lambda})_{3\times3}
	\end{bmatrix}
	\in\mathbb{R}^{4\times4}
\end{equation}

\remark{In (\ref{eq:lambda}), $\lambda_{1}$ is only used as a reference to proceed the SMA under $\vert\vert\eta_{1}\vert\vert\rightarrow0$, and at any time instant, deviation of $\eta_{1}$ off $0$ will suspend and restart the whole SMA process until $\eta_{1}$ converges again.
This implies the instrument's position (via $\eta_{1}$) is always regulated prior to the orientation (via $\eta_{2,3,4}$) due to the ``weak coupling" effect of the instrument's end-effector pose.
This is a demanding motion constraint in instrument manipulation, as the position misalignment of the end-effector caused by $\eta_{1}$ becomes significantly larger than that by $\eta_{2,3,4}$, also thanks to the ``weak coupling" effect.
(see Appendix A for mathematical explanations how ``weak coupling" appears on a wristed robotic instrument).
}

\begin{figure}[t]
\centering
\includegraphics[width=0.45\textwidth,keepaspectratio]{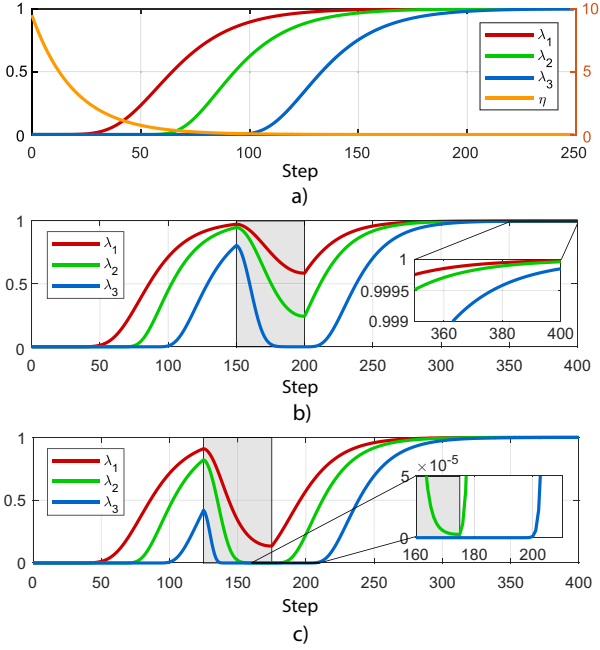}
\caption{Evolution of $\boldsymbol{\lambda}$ subject to $\vert\vert\eta\vert\vert$, where a) denotes an undisturbed evolution with exponentially descending $\vert\vert\eta\vert\vert$; b) evolution upon slight increase of $\vert\vert\eta\vert\vert$ during $t\in[150,~200]$, and c) evolution upon large increase of $\vert\vert\eta\vert\vert$ during $t\in[125,~175]$.
Note that the proximal $\lambda$ stays greater then the distal one to enforce sequential coordination.}
\label{fig:Lambda}
\end{figure}

One main advantage of introducing SMA to the DS-guided instrument manipulation is that the whole-body robot trajectory subject to (\ref{eq:ds-ds}) becomes always feasible.
This is led by (\ref{eq:lambda}) which enforces the regulation of a specific $\eta_{i},~i>1$ to be always accompanied by a settling $\eta_{i-1}$.
This further transforms the state space representation (\ref{eq:eta}) into:
\begin{equation} \label{eq:sma-state}
	    \eta_{i} =  (\textbf{C}_{i+2}{^{i+2}\textbf{R}}_{i+3}(q_{i+3})\overline{\boldsymbol{\xi}}_{i-1})^{\intercal}{\textbf{e}_{i}}\impliedby\lambda_{i-1}\rightarrow1
\end{equation}
with $i=1,2,3$, where $\textbf{C}_{3,4,5}$ are matrices whose variations are negligible due to $\lambda_{1,2,3}\rightarrow1$, $\textbf{e}_{1} = \vec{x}_{g},~~\textbf{e}_{2} = \vec{y}_{g},~~\textbf{e}_{3} = \vec{z}_{g}$.
Note that, the regulation of each $\eta$ is now dominated by only one robot joint.
Then, based on the property revealed in Proposition 2 and the controller (\ref{eq:ds-u}), the process of SMA is transformed to a sequence of globally stable control subsystems coordinated by $\textbf{L}$.
As long as $\mathcal{F}_{g}$ is reachable, $\mathcal{H}$ must be reachable and thus $\boldsymbol{\eta}=\textbf{0}$ could be reached by using (\ref{eq:ds-u}).
Examples of state evolution upon SMA and its performance when dealing with disturbances from $\boldsymbol{\eta}$ are demonstrated in Fig. \ref{fig:Lambda}.

Another explicit advantage of SMA is that the motion control of the robot's end-effector and body parts does not rely on computing inverse kinematics and thus allows a differentiable system, which could guarantee efficiency and smoothness when dealing with complicated motions.
Recall the goal position of $\textbf{x}_{s}$ is coarsely computed as $\tilde{\textbf{x}}_{s_{g}}$ in (\ref{eq:x_sg}).
During the manipulation, movements of the distal joints (e.g. $q_{5,6}$) might change $\textbf{x}_{e}$ and also $\tilde{\textbf{x}}_{s_{g}}$, which generates disturbance to $\eta_{1}$ and also the whole SMA process.
However, this facilitates $\tilde{\textbf{x}}_{s_{g}}\rightarrow\textbf{x}_{s_{g}}$ as "self-propagating" goal pose reaching process among $\mathcal{H}$, as $\eta_{2,3,4}$ are strictly bounded and globally converges to 0.
This together with the ``weak coupling" effect of the wrist kinematics leads $\textbf{x}_{t}$ to a bounded disturbance which does not affect our control system.


\subsubsection{5.1.3. Collision Avoidance \\}

\hl{In this part, we briefly introduce our framework to deal with obstacle avoidance.
In robotic surgery, obstacle avoidance mainly solves two scenarios: tool-tool collision when performing dual-arm surgical tasks within a shared workspace, and tool-tissue collision to prevent trauma to surrounding tissues.
As the obstacles might be described as a single obstacle $\mathcal{O}$ or a set of obstacles $\mathcal{O}=\{\mathcal{O}_{1},\mathcal{O}_{2},\ldots,\mathcal{O}_{n}\}$ from image-based measurements, without loss of generality, we start by considering a single obstacle identity $\mathcal{O}$ as a sphere (which could be easily extended to lines or surfaces) with 3D position $\mathbf{x}_{o}$.
From robot kinematics and eye-to-hand data, the closest point $\mathbf{x}_{clo}$ on the robot skeleton relative the obstacle, and its local Jacobian matrix $\mathbf{J}_{clo}\in\mathbb{R}^{3\times3}$ can be computed.
Note that the position of $\mathbf{x}_{clo}$ on the robot might be changing upon robot motions.
Then, by giving a minimum tolerant distance $r_{o}$ when avoiding collision (i.e. the radius of the sphere), the avoidance could be done by finding the tangent direction for the sphere $\mathcal{O}$ from $\mathbf{x}_{clo}$
There are infinite directions to satisfy such, but one could choose the closest one to $\mathbf{x}_{s_{g}}-\mathbf{x}_{clo}$ to reduce trajectory length.
Thus, an instant target position $\mathbf{x}_{g,o}$ for $\mathbf{x}_{clo}$ could be derived by letting}
\begin{equation} \label{eq:oa-1}
	    \mathbf{x}_{g,clo} = \mathbf{x}_{clo} + a_{e}\mathbf{e}_{g}
\end{equation}
\hl{where $\mathbf{e}_{g}\in\mathbb{R}^{3}$ is a unit vector that denotes the selected avoiding direction, $a_{e}$ is a scalar denoting the target motion speed.
This indicates that we directly seek for obstacle avoidance by deriving the escaping velocity to be followed by the $\mathbf{x}_{clo}$.
This allows that once $\mathbf{x}_{clo}$ goes through $\mathbf{x}_{g,clo}$, the tangent direction changes and then varies $\mathbf{x}_{g,clo}$ upon (\ref{eq:oa-1}) such that the collision avoidance motion continues.
Meanwhile, we monitor whether the obstacle still obstructs the robot to reach the target by computing the distance between $\mathbf{x}_{obs}$ and the line formed by $\mathbf{x}_{g_{s}}-\mathbf{x}_{g,clo}$.
Finally, we decide whether the robot completes obstacle avoidance by the following}
\begin{equation} \label{eq:oa-2}
\mathcal{X}_{g} = 
	\begin{cases}
        \mathcal{X}_{o}, & f_{o}(\mathcal{X},\mathcal{X}_{p},\mathbf{p}_{o},r_{o})\leq0 \\
        \mathcal{X}_{p}, & f_{o}(\mathcal{X},\mathcal{X}_{p},\mathbf{p}_{o},r_{o})>0
	\end{cases}
\end{equation}
\hl{where $f_{o}(\cdot)$ denotes the scalar function that compute the aforementioned point-to-line distance subtracted by $r_{o}$.
$\mathcal{X}_{o}$ is the end-effector state for collision avoidance constructed by $\mathbf{x}_{g}$ with unchanged orientation, as the local Jacobian only regulates the 3D position of $\mathbf{x}_{clo}$.
For practicality, to avoid abrupt change of the instant target state led by (\ref{eq:oa-2}), one could apply linear fusion of two states to achieve smooth state transition.}

\subsection{5.2. Integrated Planning and Control}

Apart from instrument manipulation, a majority of surgical instrument tasks (e.g. dissection and suturing) involve reaching and contact a specific target.
\hl{The end-effector is normally manipulated to a proper candidate pose first (i.e. the tool-centric action) and then proceed to the final contact phase (i.e. the target-centric action) to could avoid unnecessary collisions or hazardous movements} \shortcite{cao1996task,jun2012robotic}.
Coordination between these two actions should be based on online awareness of the tool-target situation such that it could adaptively decide whether the robot is ready for contact or requires further adjustment.
To this end, instead of regarding them as separated steps, we propose a unified strategy that integrates the tool-centric and target-centric actions to a single motion step to facilitate safer, smoother, and more efficient motion performance.

Recall the instrument's goal configuration described by frame $\mathcal{F}_{g}$, It now also needs to by used to guide the end-effector to direct contact a target besides manipulation (mostly grasping an object).
To this end, we extend $\mathcal{F}_{g}$ to two poses, the pre-contact pose $\mathcal{F}_{p}$ and a contact pose $\mathcal{F}_{c}$ to be reached by the instrument's end-effector.
It should be noted that $\mathcal{F}_{c}$ is usually computed via online sensoring feedback or assigned by a user, and $\mathcal{F}_{p}$ is computed based on $\mathcal{F}_{c}$ which should be reached in advance in order to proceed to the contact safely.
Then, we define a known end-effector trajectory $\rho(\tau(t))$ from $\mathcal{F}_{p}$ to $\mathcal{F}_{c}$ to be tracked by the end-effector to clear the final tool-target distance for finishing the contact, which also exactly defines the target-centric action.
The parameter $\tau(t)=[0,1]\in\mathbb{R}$ denotes the progress of the trajectory and is of class $\mathcal{C}^{1}$.
Thus, the appearance of $\tau$ enables a dynamic $\mathcal{F}_{g}$ and leads the differentiation of $\boldsymbol{\eta}$, currently of $\mathcal{X}(\textbf{q})$ and $\mathcal{X}_{g}(\tau)$, to the following extended form:
\begin{equation} \label{eq:gvm-ds}
        \boldsymbol{\dot{\eta}}(\mathcal{X}(\textbf{q}),\mathcal{X}_{g}(\tau))=  \frac{\partial\boldsymbol{\eta}(\cdot)}{\partial\mathcal{X}(\textbf{q})}\frac{\partial\mathcal{X}(\textbf{q})}{\partial\textbf{q}}\dot{\textbf{q}} 
        +\frac{\partial\boldsymbol{\eta}(\cdot)}{\partial\mathcal{X}_{g}(\tau)}(\mathcal{X}_{c}-\mathcal{X}_{p})\dot{\tau}
\end{equation}
whose evolution performance could be proved globally stable as of (\ref{eq:ds-ds}) by deploying the following controller:
\begin{equation} \label{eq:gvm-u}
    \begin{aligned}
        \textbf{u}= & \left(\frac{\partial\mathcal{X}(\textbf{q})}{\partial\textbf{q}}\right)^{-1}\left(\frac{\partial\boldsymbol{\eta}(\cdot)}{\partial\mathcal{X}(\textbf{q})}\right)^{-1}\\
        & ~~~~~~~~~~~~~~~\left(\frac{\partial\boldsymbol{\eta}(\cdot)}{\partial\mathcal{X}_{g}(\tau)}(\mathcal{X}_{p}-\mathcal{X}_{c})\dot{\tau}-\boldsymbol{\Gamma}\boldsymbol{\eta}(\cdot)\right)
    \end{aligned}
\end{equation}
who not only stabilizes $\boldsymbol{\eta}$ to $\boldsymbol{0}$ as in (\ref{eq:ds-ds}), but further guarantees that the robot state $\mathcal{X}$ will converge to the following 
\begin{equation} \label{eq:gvm-state}
    \lim_{t\rightarrow\infty}\mathcal{X}=\mathcal{X}_{g}=\tau\mathcal{X}_{c}+(1-\tau)\mathcal{X}_{p}
\end{equation}
which indicates that the robot state will now arrive exactly at the prescribed trajectory $\rho(\tau)$ that connects $\mathcal{X}_{p}$ and $\mathcal{X}_{c}$.
The instant $\mathcal{X}_{g}$ to be reached by the robot is determined by the value of $\tau$ whose online adjustment will specifically enable
\begin{equation} \label{eq:lambda}
    \lim_{t\rightarrow\infty}\vert\vert\boldsymbol{\eta}(\cdot)\vert\vert=\boldsymbol{0}\implies\mathcal{X} = 
    \begin{cases}
        \mathcal{X}_{c}, & \lim_{t\rightarrow\infty}\tau=\infty \\
        \mathcal{X}_{p}, & \lim_{t\rightarrow\infty}\tau=0 
    \end{cases}
\end{equation}
where the performance of regulating $\boldsymbol{\eta}$ will not be affected by the change of $\tau$, but meanwhile facilitates a self-tunable end-effector goal pose $\mathcal{X}_{g}$.
We define this as a goal-varying manipulation (GVM) problem that facilitates the versatile robot motions via prescribed trajectories and/or states, by setting $\mathcal{X}_{g}$ to a set of target states $\{\mathcal{X}_{g_{1}},\mathcal{X}_{g_{2}},\ldots,\mathcal{X}_{g_{n}}\}$ that characterize the  goal-relevant manipulation process.
Here, $\tau$ is also regarded as the goal varying parameter (GVP) and will be used to tune the robot's instant goal state adaptively based on its on-the-fly configuration $\boldsymbol{\eta}$, in order to guide the robot through specific motion patterns via a controllable process upon the task's needs.
To make GVM contribute to our case, we propose the following dynamics for $\tau$:
\begin{equation} \label{eq:tau-ds}
        \dot{\tau}(\boldsymbol{\eta}) = \gamma(2\exp(-\kappa\vert\vert\boldsymbol{\eta}\vert\vert)-1)
\end{equation}
where $\gamma$ and $\kappa$ are the tuning parameters of the dynamical performance.
Particularly, the dynamics (\ref{eq:tau-ds}) owns the following properties that contribute to our application in surgical task autonomy.
\begin{itemize}[leftmargin=*]
    \item \textbf{Property 1}. The alignment of $\mathcal{X}$ to $\mathcal{X}_{g}$, characterized by $\vert\vert\boldsymbol{\eta}\vert\vert$, tunes both the magnitude and the direction of $\tau$:
    \begin{equation} \label{eq:tau-ds}
            \dot{\tau} = 
            \begin{cases}
               -\gamma, & \lim_{t\rightarrow\infty}\vert\vert\boldsymbol{\eta}\vert\vert=\infty \\
                \gamma, & \lim_{t\rightarrow\infty}\vert\vert\boldsymbol{\eta}\vert\vert=0 
            \end{cases}
    \end{equation}
    where alignment of the robot to its instant goal state will elevate $\tau$ upon the tuning step up to $\gamma$, which carries forward the prescribed GVM process, while misalignment (with $\eta\nrightarrow0$ renders $\dot{\tau}<0$ which rewinds the GVM process back as $\dot{\tau}<0$ drags $\tau$ back to zero.
    This enables the manipulation process to be regulated in an adaptive and bidirectional manner guided by $\tau$.

    \item \textbf{Property 2}. Based on Property 1, $\tau$ will reach $1$ within finite time.
    This can be proved that for any instant $\boldsymbol{\eta}$, $\dot{\tau}>0$ holds once
    \begin{equation} \label{eq:tau-finite}
        \boldsymbol{\eta}(T_{1})<-\frac{1}{\kappa}\ln \frac{1}{2}
    \end{equation}
    where $T$ is the time instant when $\tau$ starts to elevate, and obviously $T<\infty$ as $\lim_{t\rightarrow\infty}\boldsymbol{\eta}=0$.
    Then, since $\dot{\tau}$ is monotonically increasing, $\dot{\tau}>0$ will hold and further lead to
    \begin{equation} \label{eq:tau-1}
        \int_{T}^{T+T'}\dot{\tau}dt>1
    \end{equation}
    where $T'<\infty$ as well if we further saturate $\tau$ in discrete time as
    \begin{equation} \label{eq:tau-finite}
            \tau(t_{k+1}) = 
            \begin{cases}
               1, & \tau(t_{k})+\dot{\tau}(t_{k})\Delta t>1 \\
                0, & \tau(t_{k})+\dot{\tau}(t_{k})\Delta t<0
            \end{cases}
    \end{equation}
     to enforce $\tau\in[0,1]$.
     This property facilitates $\mathcal{X}_{g}$ to land on the final goal state (thanks to $\tau$ (\ref{eq:tau-ds})) prior to $\mathcal{X}\rightarrow\mathcal{X}_{g}$ to ensure efficient and accurate guidance towards the target.
     
    \item \textbf{Property 3}. The boundary conditions of $\dot{\tau}$ subject to (\ref{eq:tau-ds}) evaluated with respect to $\boldsymbol{\eta}$ are:
    \begin{equation} \label{eq:tau-bound}
        \left.\frac{\partial\dot{\tau}}{\partial\boldsymbol{\eta}}\right\vert_{\boldsymbol{\eta}\rightarrow\infty}\rightarrow0,~~~~~~
        \left.\frac{\partial\dot{\tau}}{\partial\boldsymbol{\eta}}\right\vert_{\boldsymbol{\eta}=0}=-2\kappa\gamma
    \end{equation}
    which indicates that the tuning of $\tau$ is sensitive to the change of $\boldsymbol{\eta}$ around the alignment situation (i.e. $\boldsymbol{\eta}\rightarrow0$), and becomes relaxed under misalignment with $\boldsymbol{\eta}\rightarrow\infty$.
    This will make $\tau$ rapidly descend away from 1 once encountering misalignment between $\mathcal{X}$ and $\mathcal{X}_{g}$.
    
\end{itemize}
During the adjustment of $\tau$, we regard $\boldsymbol{\eta}$ as an external input whose regulation is solely subject to (\ref{eq:gvm-u}).
Note that, the introduction of GVM using $\tau$ on the basis of DS-guided manipulation in (\ref{eq:gvm-ds}) further integrates the trajectory planning and the motion control of versatile robot manipulation steps to a unified solver.
Selection of individual $\mathcal{X}_{g}$ and their connected trajectory $\rho$ will enable versatile motion characteristics, which could be either pre-defined or solved on-the-fly.

\begin{figure}[t]
\centering
\includegraphics[width=0.48\textwidth,keepaspectratio]{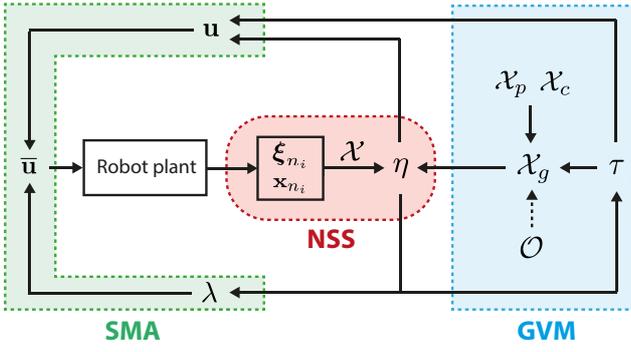}
\caption{\hl{The schematic diagram of our integrated planning and control framework using NSS $\boldsymbol{\eta}$ to represent and control the robot instead of end-effector pose.
GVM serves as an adaptive planning policy that provides active guidance using the goal state $\mathcal{X}_{g}$ via $\tau$, which is reactive to obstacle avoidance.}}
\label{fig:ParameterFramework}
\end{figure}

The above GVM framework could be used to automate typical instrument motions in robotic surgery.
First, it could solve delicate motions required for target contact.
We can simply define $\mathcal{X}_{p}$ whose orientation is identical to that of $\mathcal{F}_{c}$ but with a constant position difference $d$ along the tool's pointing direction, with
\begin{equation} \label{eq:d}
    d > l_{w} + l_{t}
\end{equation}
to ensure the tool-centric action keeps the manipulation clear from the target before it reaches $\mathcal{X}_{p}$.
It degenerates $\rho$ to a line segment $\overrightarrow{\mathcal{O}_{p}\mathcal{O}_{c}}$, such that the robot will first reach $\mathcal{X}_{p}$ with the orientation already suitable for contact, and then clears the remaining distance $d$ upon $\tau\rightarrow1$ to finally reach $\mathcal{X}_{c}$.
Here, the tool-centric action (reaching $\mathcal{X}_{p}$) and target-centric action (reaching $\mathcal{X}_{c}$) are independent but smoothly coordinated.
Second, GVM can be applied for tracking trajectories required by specific task-relevant motions (e.g. needle insertion path must follow its own circular curvature to minimize tissue trauma).
Here, $\rho(\tau)$ denotes a general path which starts from $\mathcal{X}_{g_{1}}$ and ends at $\mathcal{X}_{g_{2}}$.
The GVP $\tau$ will attract the robot to enter $\mathcal{X}_{g_{1}}$ and then guide $\mathcal{X}$ to follow $\rho(\tau)$ until $\mathcal{X}_{g_{2}}$ is reached.
Due to $\dot{\tau}$ in (\ref{eq:gvm-ds}) and Property 3, $\tau$ will smoothly elevates to $1$ while ensuring that $\mathcal{X}$ could catch up the varying $\mathcal{X}_{g}$.
The overall model of the integrated planning and control framework is shown in Fig. \ref{fig:ParameterFramework}.


\begin{figure}[t]
\centering
\includegraphics[width=0.33\textwidth,keepaspectratio]{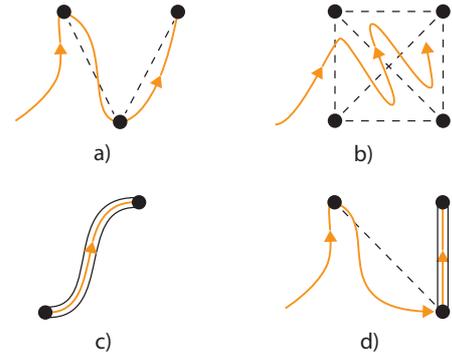}
\caption{Conceptual illustration of different types of motion behavior achievable by using the GVM structure. The orange curve indicates the guided motion by the dynamic $\mathcal{X}_{g}$ through different types of motion (black dots as the via states, dashline as free manipulation motion, parallel solid lines as motion with determined path).
}
\label{fig:GVM}
\end{figure}

We briefly emphasize the superiority of our architecture from a combined technical and practical perspective that contribute to surgical applications.
The introduction of NSS to represent the robot's present-to-goal situation instead of the traditional coupled end-effector pose.
The global asymptotic stability of (\ref{eq:ds-ds}) and (\ref{eq:gvm-ds}) guarantees that the robot can be stably maneuvered to any arbitrary feasible state with smooth positions and velocities.
Incorporating SMA then provides versatile planning of the robot's goal-reaching trajectory as well.
They solve the issues of ``the robot could reach a goal'' and ``the robot knows how to reach a goal'' simultaneously using our framework via a mathematically determined solver without prior knowledge or online iterations.
GVM then plays a role to solve ``the robot knows when to act'' which picks up the last components required for a motion-level architecture.
Its capability to transfer robot motion through different types of motions is illustrated in Fig. \ref{fig:GVM}.

\hl{For surgical applicability, GVM allows motion constraints applied to the in-process trajectory, which makes it powerful to solve delicate target contact motions, where a constrained path is required to contact tissue via specified poses to avoid unnecessary tissue trauma.
Particularly, the tool-centric action guarantees avoidance of premature contact to the target, and the instrument will not approach the target (at $\mathcal{X}_{c}$) before it reaches $\mathcal{X}_{p}$ (s.t. $\boldsymbol{\eta}\rightarrow0$).}
This could avoid entanglement of the tool to the target before the final phase of manipulation (i.e. $\mathcal{X}_{g}\rightarrow\mathcal{X}_{c}$) is settled.
In addition, as the tuning of GVP $\tau$ is smooth and bounded, the robot actions can be transferred smoothly as well.

\section{6. Task-Level Autonomy Architecture}

In this section, we extend our approach to task-level automation such that it could complete the whole pipeline of one or more surgical tasks.
The key issue is to further define a powerful motion descriptor that reveals the mutual behavior among individual motion steps while tolerating their underlying differences.
GVM has enabled safe target contact by constraining the instrument motion delicately during the final motion phase.
However, how to guide the instrument for clearance of the target or interact via specific path are similar but different steps to be considered.

\subsection{6.1. Surgical Motion Primitive (SMP)}

To systematically entail these motion details, we propose the notion of surgical motion primitive (SMP) based on GVM as a generalized model to define surgery-related instrument motions.
The SMP characterizes a single instrument manipulation step into three progressive phases:
\begin{itemize}[leftmargin=*]
    \item \textbf{Phase I (Backward guidance)}: The robot is guided via evolution of GVM through $\tau:1\rightarrow0$, which is used for target-centric action to escape the instrument escapes from an already contacting object.
    \item \textbf{Phase II (Free control)}: The robot is guided via evolution of $\tau=0$ without GVM, which is used for tool-centric action to reach the instrument to a target state without contacting any target objects.
    \item \textbf{Phase III (Forward guidance)}: The robot is guided via evolution of GVM through $\tau:0\rightarrow1$, which is used for target-centric action or trajectory tracking, where the instrument approaches and contacts a new target object or follows specific path.
\end{itemize}
The current setting of our GVM can only achieve consecutive execution across Phase II and Phase III.
Thus, we must modify GVM such that it is applicable to the whole SMP.
In this regard, we extend the validity of $\tau$ to Phase I such that it must retract the end-effector from the target to a safe position prior to Phase II.
We define the initial pose as $\mathcal{X}_{c_{0}}$ where contact still exists, and the retracted state as $\mathcal{X}_{p_{0}}$, which is also the ending state for Phase I.
Without loss of generality, we maintain the relationship between them as that of $\mathcal{X}_{c}$ and $\mathcal{X}_{p}$, and their in-between trajectory $\rho$ remains a line segment and should be performed rapid by the instrument to avoid unnecessary contact.
The GVP $\tau$ is also set to $1$ at $\mathcal{X}_{c_{0}}$ and $0$ at $\mathcal{X}_{p_{0}}$.
Thus, during Phase I, we need to enforce $\dot{\tau}=-c$ (with $c$ being a positive scalar) to lead the instrument to $\mathcal{X}_{p_{0}}$ subject to $\tau:1\rightarrow0$, which exactly reverses the process of Phase III.
To smoothly connect Phase I to Phase II, we equip $\tau$ with the following properties 
\begin{equation} \label{eq:}
    \mathcal{X}_{g} = 
        \begin{cases}
           \mathcal{X}_{p_{0}},  & \epsilon<\tau\leq1  \\
           \mathcal{X}_{p},  & 0\leq\tau< \epsilon
        \end{cases}
\end{equation}
where $\epsilon$ is the threshold that coordinates Phase I and Phase II.
Once $\mathcal{X}_{g}$ switches to $\mathcal{X}_{p}$, the dynamics of $\tau$ is switched by to (\ref{eq:tau-ds}) to regulate subsequent SMP phases.

\subsection{6.2. Modes of Behavior (MoB)}

\begin{figure*}[t]
\centering
\includegraphics[width=0.82\textwidth,keepaspectratio]{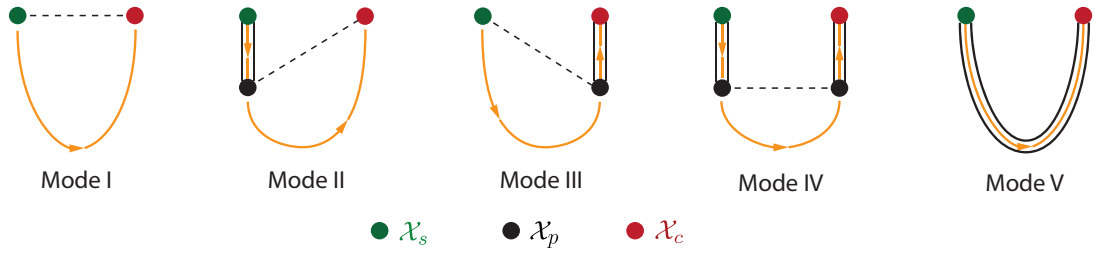}
\caption{The five modes of behaviors (MoBs) proposed based on our planning and control framework to define individual motion steps in various tasks. Mode I: Free reaching, Mode II: Constrained retraction with free reaching, Mode III: Constrained reaching, Mode IV: Constrained retraction and reaching, Mode V: Trajectory tracking.}
\label{fig:MoB}
\end{figure*}

\begin{figure*}[t]
\centering
\includegraphics[width=0.9\textwidth,keepaspectratio]{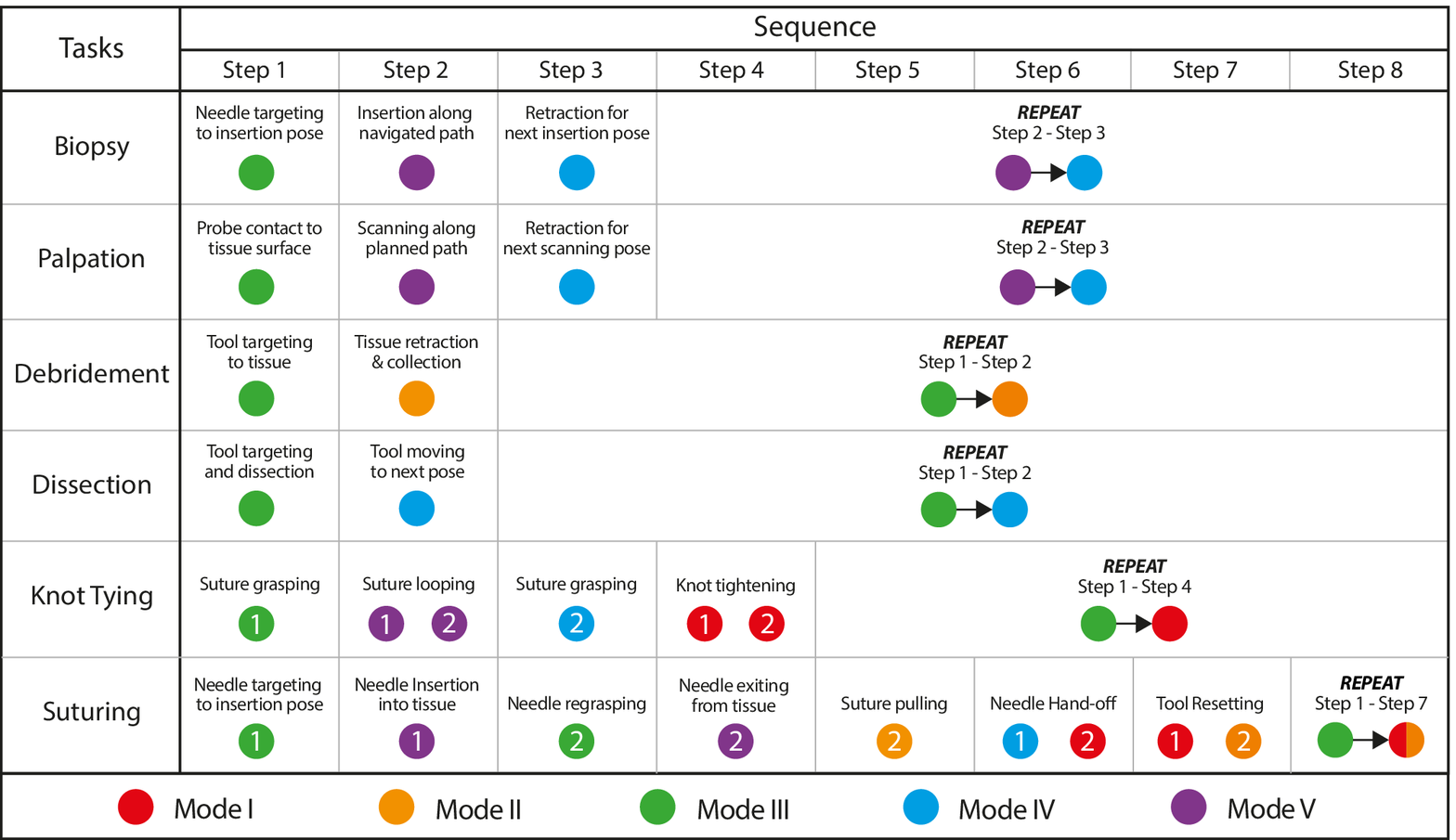}
\caption{Demonstration of complete pipelines of various independent surgical tasks which could be fully formulated by our framework using the five MoBs for tasks automation.}
\label{fig:MoBTasks}
\end{figure*}

Finally, we utilize SMP to define and automate the entire pipeline of instrument motion sequences for different surgical tasks.
Although we define three consecutive phases that form the single SMP, some instrument motion steps might only involve one or two of them.
For example, an off-target manipulation can be fully performed by initiating only Phase II.
Retracting the instrument from an object to a stand-by contact-free target could be done by using Phase I and Phase II.
To make SMP customizable to such motion differences, we combine different motion phases and generate the following five typical modes with their application scenarios:
\begin{itemize}[leftmargin=*]
    \item \textbf{Mode I (Phase II)}: The instrument motions is freely adjusted via DS+SMA with no additional workspace constraints and thus no GVM ($\tau=0$).
    \item \textbf{Mode II (Phase I + II)}: The instrument leaves a contacted object under constrained path to a off-contact state guided via GVM with $\tau:1\rightarrow0$.
    \item \textbf{Mode III (Phase II + III)}: The instrument reaches and contacts a target from an off-contact state (e.g. grasping a tissue/needle) guided via GVM with $\tau:0\rightarrow1$.
    \item \textbf{Mode IV (Phase I + II + III)}: The instrument retracts from the previous object via constrained path ($\tau:1\rightarrow0$), adjusts its pose ($\tau=0$), and then reaches and contacts a new target object, with $\tau:0\rightarrow1$.
    \item \textbf{Mode V (Phase III)}: The instrument is guided through a specific path with determined initial/final state $\tau:0\rightarrow1$.
\end{itemize}
They are also defined as the MoBs, as they exhibit own motion constraints while sharing the identical control architecture, and are available to each SMP (see Fig. \ref{fig:MoB} for conceptual illustration).
To achieve task-level autonomy, we first utilize a chain of SMPs that govern each motion step in the surgical task to construct the whole pipeline.
Then, we equip each SMP (or each motion step) with a specific MoB according to its prescribed functionality.
As the guidelines for surgical tasks are already mature in clinical practice, the MoBs could be preset for each task such that it could successfully follow the procedure.
In Fig. \ref{fig:MoBTasks}, we demonstrate the applicability of our framework to characterize many different existing surgical tasks by selecting different SMPs and MoBs, based on calinically adopted surgical guidelines \shortcite{cao1996task}.
The validity covers not only intracorporeal MIS procedures (e.g. tissue dissection, wound suturing, knot tying, etc.) but also other procedures like robotic palpation, biopsy, etc., which is generalized framework compared to existing task-specific approaches.
Although each elementary motion has different constraints, they could be set by five modes of behavior to form a pipeline.
\hl{For example, the specification for dual-arm suturing could be easily interpreted as 3-5-0-0-0-4-1 for PSM 1 and 0-0-3-5-2-1-2 for PSM 2 (where 0 means staying idle in this step).
This could all be preset in the framework as long as the task type to be automated is determined, which is very easy to specify.}
We will show later in our experiments that, the ability of coordinating motion execution process adapting to the online robot-environment situation is critical to improve the success rate and reliability of task autonomy.

\section{7. Simulations \& Results}

In this section, we present our simulation set up and results to demonstrate the motion-level performance of robotic instrument manipulation.
The simulations will be conducted to reveal the following two main aspects of our approach: How the robot motions are actuated to deal with delicate and drastic manipulation steps, and how the strategy reacts to different tool-target configurations when attempting to perform interaction to a physical target.

\subsection{7.1. Overview}

We implement our algorithm framework on the Virtual Robot
Experimentation Platform (V-REP), which is then interfacing to Matlab R2017a (MathWorks Inc) via remote API on a Core i7 2.8 GHz system without GPU accleration.
In the simulation platform, we use the virtual model of da Vinci Research Kit (dVRK) provided by \shortcite{fontanelli2018v}.
The dVRK is a dual-arm surgical robot with two wristed robotic instruments, which are named as the Patient-Side Manipulators (PSMs).
Both two attached instruments are Large Needle Drivers (LNDs) which is the typical selection in dVSS-enabled RAMIS to perform delicate intra-corporeal procedures like suturing.
\hl{Virtual objects, either pegs or a surgical needle, involve pick-and-place operation, which is one common surgical motion type and is frequently used as a proxy for generalized surgical dexterity} \shortcite{cao1996task}.
Here, we will use them to define practical task-relevant movement steps for evaluation of instrument's motions.
The distance between the RCM of the PSMs to the peg transfer objects are selected to around 150mm which resembles the genuine robot-target configuration during RAMIS.
We do not apply dynamic motion properties and interactions throughout the simulations, as we mainly investigate the motion behavior of the instruments in 3D workspace.
The detailed parameter set up of our system is shown in Table \ref{T2}.

\subsection{7.2. Scene I: Complex Tool Manipulation}

\begin{figure}[t]
\centering
\includegraphics[width=0.47\textwidth,keepaspectratio]{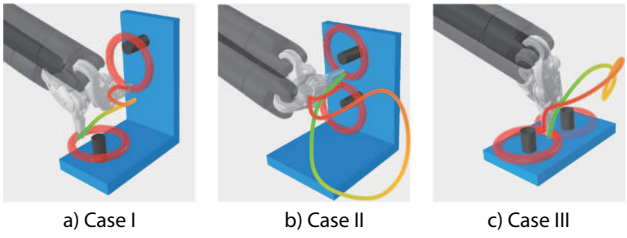}
\caption{The three cases of pose-to-pose manipulation for simulated peg transfer tasks where the instrument needs to drastically twist the wrist. The end-effector trajectories are visualized from green to red along the evolution of time.}
\label{fig:LargePoses}
\end{figure}

We first simulate a peg-transfer set up where a plastic ring is to be manipulated by the robot from one peg to another, which is commonly used in surgical skill training for novice surgeons.
The initial/final state of the pegs are usually selected that correspond to awkward robot configurations (i.e. either near-singularity or near-limit joint states).
The robot trajectory needs to experience large-range orientation adjustment in order to align the ring to the peg to finish the manipulation properly, which might be hard to plan within confined space.
To verify the validity of our framework to tackle such extreme scenarios, we assign three typical cases, as shown in Fig. \ref{fig:LargePoses}:
\begin{itemize}
	\item Case I: Transfer the ring from a horizontal peg to an adjacent vertical peg.
	\item Case II: Transfer the ring from a horizontal peg to another horizontal peg by twisting the shaft through 180 degrees.
	\item Case III: Transfer the ring from a vertical peg to another vertical peg by twisting the shaft through 180 degrees.
\end{itemize}

\begin{table}[t]
\footnotesize\sf
\caption{Parameter setting of our algorithm in simulations.\label{T2}}
\begin{tabularx}{0.47\textwidth}{r|l}
\toprule
~~~~~~~~~~~~~~~~~Parameter & Value~~~~~~~~\\
\midrule
\vspace{0.1cm}
$\Gamma$ &
$\left[ \begin{array}{ccc} 0.1 & 0 & 0 \\ 0 & 2 & 0 \\ 0 & 0 & 2 \end{array}\right]$\\
\vspace{0.1cm}
$\textbf{l}_{h}$ & $\left[\begin{array}{ccc} -1 & 0 & 0 \end{array}\right]^{\intercal}$\\
\vspace{0.1cm}
$\textbf{l}_{v}$ & $\left[\begin{array}{ccc} 0 & 1 & 0 \end{array}\right]^{\intercal}$\\
\vspace{0.1cm}
$\textbf{m}$ & $\left[\begin{array}{ccc} 0 & 0 & 1 \end{array}\right]^{\intercal}$\\
\vspace{0cm}
$k_{1}$ & $\times$ (random)\\
~$k_{2}$ & $0.2$\\
\vspace{0cm}
$k_{3}$ & $1$\\
\vspace{0cm}
$k_{4}$ & $1$\\
\vspace{0cm}
$\gamma$ & $0.05$\\
\vspace{0cm}
$\kappa$ & $0.5$\\
\vspace{0cm}
$\epsilon$ & $0.01$\\
\bottomrule
\end{tabularx}\\[0pt]

\end{table}

\begin{figure}[t]
\centering
\includegraphics[width=0.48\textwidth,keepaspectratio]{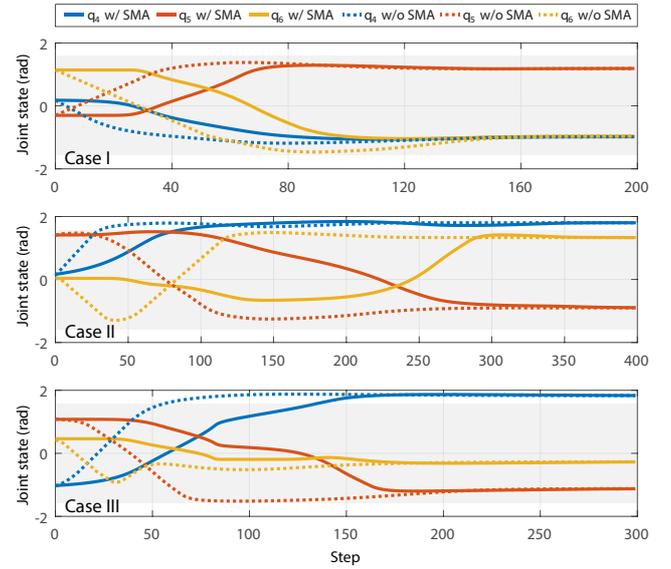}
\caption{Evolution of distal joint positions for three cases of complex manipulation.
The green areas denote the $-\frac{\pi}{2},\frac{\pi}{2}$ feasible joint range available to both $q_{5}$ and $q_{6}$, where motion control without SMA tends to overshoot the joint positions before convergence, which might enter unreachable regions under near-limit configurations.}
\label{fig:LargePosesJoint}
\end{figure}

\begin{figure}[t]
\centering
\includegraphics[width=0.48\textwidth,keepaspectratio]{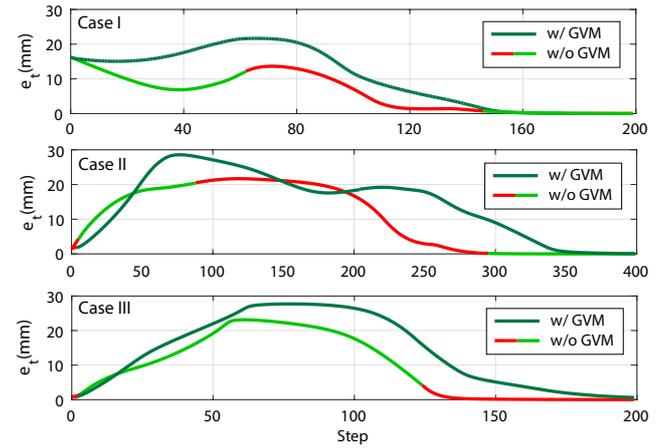}
\caption{Instrument end-effector position error during target-reaching manipulation in the three cases with and without using GVM-based guidance. The red parts indicate the time instants where target intrusion happens (i.e. $\phi<0$).}
\label{fig:LargePosesIntrusion}
\end{figure}

The target states of the robot are known and remain static at this stage of simulation.
The process is actuated using Mode III for our proposed SMP.
Particularly, the differences between initial and final positions of the distal joints $q_{w_{1}}$, $q_{w_{2}}$, $q_{t}$ in three cases are $[1.16, 1.487, 2.1]^{\intercal}$ rad, $[1.672, 2.294, 1.296]^{\intercal}$ rad and $[-2.853, 0.732, 2.2002]^{\intercal}$ rad, respectively, which indicate an average of $>100^{\circ}$ motion for each distal joint in each case.
The results show that the DS-based framework could simultaneous plan and control the instrument's motion through twisted motions and reach the target configuration in all cases.
The guaranty of global unique solution in Proposition 2 in our new robot model, the instrument smoothly reaches the target pose without being trapped in false equilibriums, which might appear in optimization-based approaches due to the nonconvex workspace manifold.
Meanwhile, the trajectory remains feasible despite the target joint positions of specific joints being adjacent to its limits.
We compare the resultant joint actions with and without deploying the SMA (as shown in Fig. \ref{fig:LargePosesJoint}.
Without SMA, the distal joint motions tend to overshoot from its final positions before convergence.
This might result in exceeding joint limits ((e.g. $q_{t}$ in Case I and II, and $q_{w_{2}}$ in Case III, note that the joint limit of $q_{w_{2}}$ and $q_{t}$ are both $\pm\pi/2$ rad, illustrated by green shaded areas).
The convergent performance is also not affected by the sequential actions.

Then, we verify the capability of GVM to inherently avoid unnecessary contact to the target during manipulation.
The main symbol of inadvertent tool-target collision is the time instant(s) that the end-effector position goes behind the target contact point, which is mathematically interpreted by the following metric:
\begin{equation} \label{eq:intrusion}
    \phi:=\langle^{g}\textbf{x}_{t}, \vec{z}_{g}\rangle < 0
\end{equation}
Guiding the end-effector through $\mathcal{x}_{p}$ and $\mathcal{x}_{c}$ by implementing GVM avoids $\phi<0$ throughout three cases (refer to the dark green curves in Fig. \ref{fig:LargePosesIntrusion}).
Without GVM, the end-effector will intrude the target area (shown in Fig. \ref{fig:LargePosesIntrusion} the red curve segments) during manipulation which could damage the tissue in real surgical procedures.
GVM allows the twisted motions to be properly adjusted first safely away from the target before reaching it.
Note that such performance is not particular to the selected cases, but could be easily guaranteed upon proper setting of $\mathcal{X}_{p}$ according to $\mathcal{X}_{c}$ in (\ref{eq:gvm-state}) could guarantee
Note also from Fig. \ref{fig:LargePosesIntrusion} that a reactive $\mathcal{X}_{g}$ does not affect the convergent performance of instrument manipulation.

\begin{figure}[t]
\centering
\includegraphics[width=0.46\textwidth,keepaspectratio]{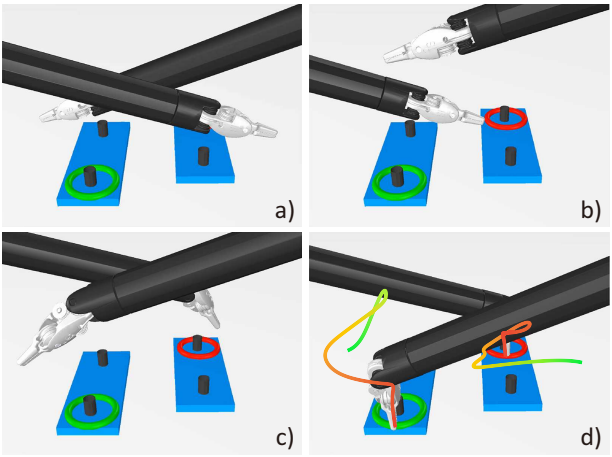}
\caption{\hl{Simulation snapshots of dual-arm manipulation for ring grasping with tool-tool collision avoidance that evolutes as a)-d). In d), the back-projected instrument end-effector positions are shown that move from green to red.}}
\label{fig:OA_illu}
\end{figure}

\begin{figure}[t]
\centering
\includegraphics[width=0.46\textwidth,keepaspectratio]{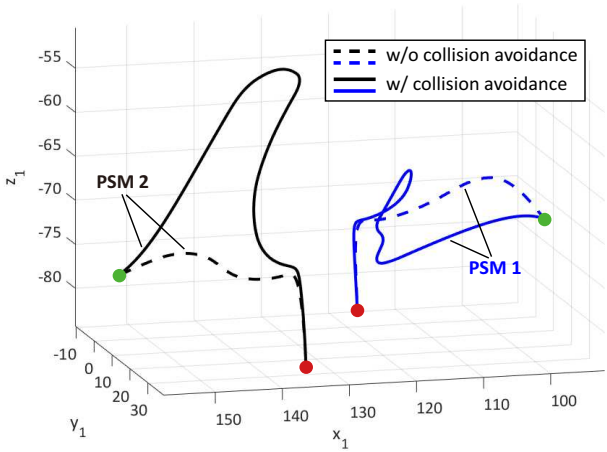}
\caption{\hl{Motion trajectories of two robotic instruments during simultaneous ring grasping in Scene II (described by the base of PSM1 in mm) with and without applying collision avoidance. The green/red dots indicate initial/final positions of two robot end-effectors, respectively.}}
\label{fig:OA_traj}
\end{figure}

\subsection{7.3. Scene II: Dual-Arm Collision Avoidance}

\hl{Then, we investigate the applicability of our algorithm to deal with collision avoidance.
Aiming for a common but challenging scenario in robotic surgery, we consider a tool-tool collision situation where two robotic instruments are manipulated to their respective targets respectively, and have to move close to each other within shared workspace.
The initial and final states of two instruments are defined such that their trajectories will make the robot skeleton collide to each other if no avoidance is actuated.}

\hl{Fig. \ref{fig:OA_illu} shows the set-up of this scene and the simulation snapshots for the dual-arm manipulation considering tool-tool collision avoidance.
Our framework allows simultaneously dual-arm collision avoidance as the planning strategy is decentralized.
The minimum tolerant distance is set to 10 mm as the tool shaft size is 8 mm, leaving the minimum tool-tool distance to be 2 mm.
Both two instruments start collision avoidance as soon as the process begins and consider the other one as the moving obstacle.
By inspecting the instant nearest obstacle point on the skeleton, two shafts move around each other and retract the distal part to prevent the whole robot from collision.
After the other instrument stays clear from its target, the instrument proceeds on the tool-centric action and target-centric action.}

\hl{Fig. \ref{fig:OA_traj} shows the resultant dual-arm trajectories.
For comparison, the trajectories for direct target reaching are shown as well.
It is clear that to avoid collision, the trajectories are more complex, but remain smooth and does not affect the subsequent target reaching process where a constrained path is required.
The collision avoidance input only applies on demand when the obstacle obstructs target reaching, and will smoothly fade once the obstacle is clear.}

\begin{figure}[t]
\centering
\includegraphics[width=0.45\textwidth,keepaspectratio]{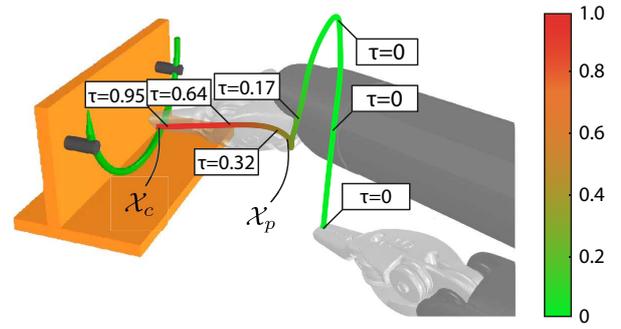}
\caption{Illustration of pre-grasp manipulation and final constrained target contact planning and control upon evolution of $\tau$ based on (\ref{eq:tau-ds}).}
\label{fig:ReachingBasicTraj2D}
\end{figure}

\subsection{7.4. Scene III: Perturbed Interaction}

\begin{figure}[t]
\centering
\includegraphics[width=0.46\textwidth,keepaspectratio]{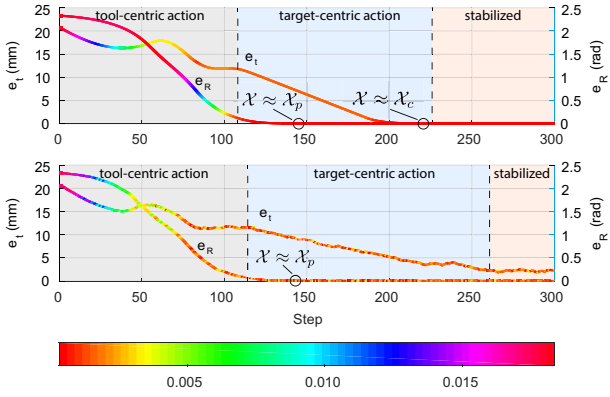}
\caption{Evolution of two examples of target reaching, shows that orientation is alignment in tool-centric action phase, then comes to translation to eliminate the remaining distance to avoid premature collision.}
\label{fig:ReachingRT}
\end{figure}

In this subsection, we validate the motion performance of the  instrument when reacting to different target situations led by the GVM.
A typical action during surgery is to manipulate the instrument's distal tool from a safe idle configuration to contacting a target object, usually for grasping purposes.
The contact must be made via a proper reaching pose to ensure safe tool-target interaction.
We use a half-circle (or 1/2) surgical needle as a target, where the robot aims to grasp the needle body at its two-thirds curvature point as a standard needle pick-up procedure.
The needle is attached to a support to visually emphasize its pose change during simulation, which is also known to the robot's base.
Again, we apply Mode III motion behavior to the instrument, where $\mathcal{X}_{g}$ moves between $\mathcal{X}_{p}$ and $\mathcal{X}_{c}$ subject to on-the-fly regulation of $\tau$.
$\mathcal{X}_{p}$ is assigned with the same orientation as of $\mathcal{X}_{c}$ but with a negative z-axis offset position from the grasping point.
Fig. \ref{fig:ReachingBasicTraj2D} illustrates the tool reaching process of the instrument from its idle position $[-116.04, 16.85, -56.13]^{\intercal}$ mm to the ideal needle grasping point $[-135.68, 18.60, -61.17]^{\intercal}$ mm.
The trajectory of the end-effector indicates that the robot performs tool-centric action that guides the tool to the pre-grasp pose first with $\tau$ maintaining close to 0.
While $\mathcal{X}_{g}$ is nearly aligned with $\mathcal{X}_{p}$, the GVP $\tau$ adaptively transits from $0$ to $1$ to further allow the instrument to eliminate the remaining distance for contact, i.e. proceeding to target-centric action.
Note that, during such action, the orientation of the distal tool remains unchanged despite the varying $\mathcal{X}_{g}$ at the final phase of contact motion, which helps achieve a proper tool-target contact configuration.
It is also shown in Fig. \ref{fig:ReachingRT}, that the regulation of the end-effector's orientation is already settling (with the robot approaching $\mathcal{X}_{p}$) before the target centric action begins (with the robot approaching $\mathcal{X}_{c}$).
This allows the tool to be "well-prepared" in advance to delicately reach the object, which will be shown later in our experiments to improve task-level reliability.

We next study how GVM will react to a target under different situations.
The target pose of the robot might be perturbed, where the disturbances could be from sensoring noises or physiological motions during surgical procedures.
We add uniform random noises to the target 3D position with the magnitude of 0.003 mm for all axial directions and is deployed throughout the process.
The orientation remains unchanged at this stage.
The comparison of the target reaching motion performance with and without applying noises is shown in Fig. \ref{fig:ReachingRT}.
It is clear that the perturbed setting results in not only longer convergent time of the tool's position $e_{\textbf{t}}$, but also remaining a stabilized position difference (denoted by $e_{\textbf{t},s}$) from the ideal contact point.
This results from the untrackable noise which makes the robot controller hard to converge $\eta$ sufficiently to $0$, and thus prevent $\tau$ from settling to 1, although the orientation convergence is not affected.
This indicates the the noisy target could lead to an "inconfident" targeting attempt by our planning and control framework.

\begin{figure}[t]
\centering
\includegraphics[width=0.47\textwidth,keepaspectratio]{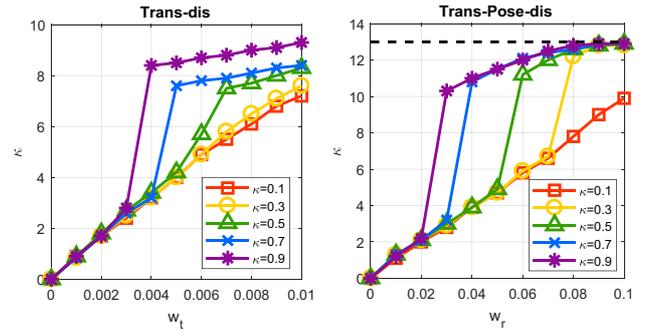}
\caption{Stabilized end-effector distance from the final contact state $\mathcal{X}_{c}$ under different tuning gain $\kappa$ and different types of noises.}
\label{fig:NoiseComp}
\end{figure}

To further study the robot motion behavior using our system to deal with target disturbances, we apply different noises to the identical set up and further obtain the relationship between the $e_{\textbf{t}}$ and the noise configurations, shown in Fig. \ref{fig:NoiseComp}.
Obviously, larger noises indicate larger stabilized position error $e_{\textbf{t},s}$ from which the relationship is relatively linear.
Meanwhile, selecting smaller gain $\kappa$ that tunes the reaction of  $\dot{\tau}$ to $\boldsymbol{\eta}$ will lead to decreased $e_{\textbf{t},s}$.
This is because a smaller $\kappa$ relaxes the condition of position alignment between $\mathcal{X}$ to $\mathcal{X}_{g}$ and ease $\tau$ to move to 1.
When the noise magnitude is elevated to a certain turning point, the stabilized position rapidly increases but remain linear afterwards.
This is caused by the $\tau$ rewinding from nearly 1 to $\ll$1 due to the noise-induced misalignment of the tool to the target (or the unsettled $\boldsymbol{\eta}$.
$\mathcal{X}_{g}$ then returns to $\mathcal{X}_{p}$ to make easier for the robot to reach, and thus automatically creating a new balance.
When applying both position and orientation noises to the target (with 0.001 mm magnitude for position and tunable for orientation, both random uniform noises).
As the change of $\mathcal{X}_{g}$'s orientation provides extra disturbance to the position of $\mathcal{X}_{p}$, the turning point of $e_{\textbf{t},s}$ upon increasing $\kappa$ comes earlier, while the level-off $e_{\textbf{t},s}$ is also higher than that without orientation noise.
However, the maximum $e_{\textbf{t},s}$ under different configuration will not exceed $d$ in (\ref{eq:d}), as in the worst case, $\tau$ will only drop to 0.
The above performance reflects that the disturbance of the target state could significantly change the motion behavior of the instrument when attempting to reach the target.
Target perturbation will make the strategy cautious when performing the target-centric action, and might even abandon it if the disturbance is severe enough due to the impossibility to settle $\boldsymbol{\eta}$ to a.
The instrument will still settle to $\mathcal{X}_{g}$ with a contact-ready orientation upon tool-centric action, but then keeps a (preset) safe distance from the target.
If the noise magnitude becomes small, the robot controller is still be able to converge $\boldsymbol{\eta}$ close to 0 such that $\tau$ maintains positive as in (\ref{eq:tau-ds}).
The strategy is then "confident" enough to guide the end-effector to the $\mathcal{X}_{c}$, or the final contact pose (e.g. $d<2$ mm or so).
Note that among all the above cases, a stabilized $\mathcal{X}$ and $\mathcal{X}_{g}$ is always achieved.
The performance of the GVM using the proposed framework remains stable regardless of the noises being added.

\begin{figure*}[t]
\centering
\includegraphics[width=0.98\textwidth,keepaspectratio]{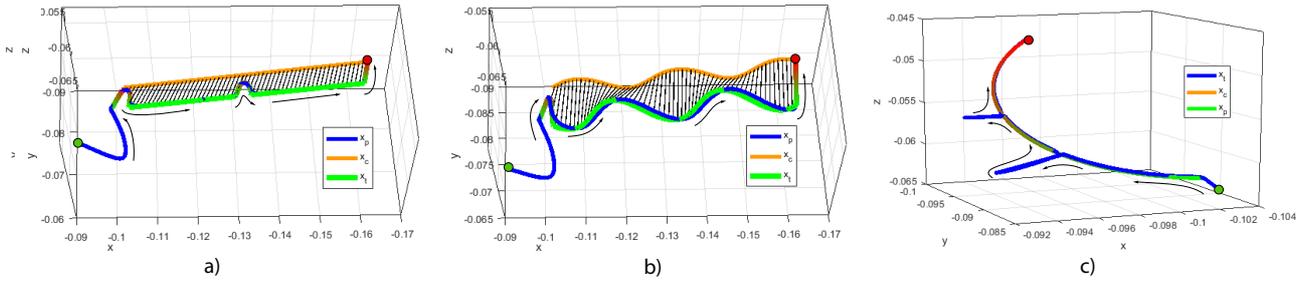}
\caption{Resultant trajectories of the instrument in all three cases in Simulation Scene III. The orange and blue curve represents the target and the instrument's end-effector position, respectively. The curve with changing color between green and red is the trajectory of the pregrasp point, or the position evolution corresponding to $\mathcal{X}_{p}$.}
\label{fig:ReachingGVM}
\end{figure*}

\subsection{7.5. Scene IV: Target-Varying Interaction}

Finally, we conduct simulation based on Scene II to study how our framework could deal with a moving target.
The instrument might reach a target which is being moved by another instrument (e.g. needle hand-off in wound suturing) or need to follow specific trajectories, such as needle insertion into tissue and palpation through tissue surfaces.
The target is either moving or to be moved after being grasped.
Here, we propose the following three typical cases that covers different target motion types to further evaluate the applicability of our framework to different manipulation strategies:
\begin{itemize}
	\item Reaching a intermittently moving target along a linear trajectory with unchanged orientation.
	\item Reaching a continuously moving target along a pose-varying trajectory (moving position and orientation).
	\item Tracking an arc-shape trajectory (moving position and orientation)	
\end{itemize}
The first and two cases correspond to decentralized dual-arm coordination for target hand-off are set.
In the first case, the final contact position $\textbf{x}_{c}$ (or the needle) moves from $[-102.59	-86.77	-62.46]^{\intercal}$ mm to $[-163.49 -85.99 -58.49]^{\intercal}$ mm through a straight line but halts at the midpoint for 100 steps.
The movement starts at $t=180$ and the overall travel distance is around $60$ mm.
The target orientation is $[2.7013    0.5169   -1.8761]$ and remains unchanged.
The translating velocity of the needle is set to $0.1$ mm per step.
In Fig. \ref{fig:ReachingGVM}a), the end-effector trajectory is first adjusted for tool-centric action for target orientation alignment, and then proceed to target-centric action while $\tau$ is moving to 1.
However, after the needle starts to move (in orange), the misalignment between the tool and the target makes contact process immediately suspended, where the instrument position rapidly moves back to $\mathcal{X}_{p}$ and track the needle's pose with a stabilized distance $d$.
While the needle temporarily stops, the instrument attempts to reach the target again, and is suspended again as the needle moves.
Finally, after the needle comes to its final stop, the instrument manages to reach the final contact point as $\tau$ eventually settles at 1.

Case II is similar to Case I but with 6-DoF pose change over time.
By reserving the translation motion of the needle, we add rotary motion of the needle with respect to its own frame, i.e. $0.15*\text{sin}0.025t$ rad along y-axis and $0.25*\text{sin}0.025t$ along z-axis.
The motion starts at $t=150$ and continuously lasts for $\delta t=600$.
As shown in Fig. \ref{fig:ReachingGVM}b), the end-effector manages to track the moving $\mathcal{X}_{g}$ induced by $\mathcal{X}_{c}$ without contacting the needle.
When the motion disappears, the tool is finally guided to the contact point where the needle could be directly grasped.

In Case III, we show how GVM contribute to stable trajectory tracking of the instrument.
We first define an arc-shaped trajectory whose distance to the rotation center is $15$ mm, and the total equivalent rotation range is $\frac{2}{3}\pi$.
Such needle insertion path is commonly used in MIS to minimize tissue trauma during penetration.
The GVP $\tau$ becomes the parametric value that guides $\mathcal{X}_{g}$ along the trajectory.
The robot will move to the starting point of the trajectory first and then follow the trajectory as $\tau$ elevates to 1.
While the robot output meets disturbances, i.e. deliberate deviation of the tool from the prescribed path, the value of $\tau$ will not move until the controller guides the tool back on where it deviates from (illustrated in Fig. \ref{fig:ReachingGVM}c)).


\subsection{7.6. Summary of Simulations}

The simulation results demonstrate the motion-level behavior of our robot planning and control framework in different scenes.
Overall, the GVM method is validated to be capable to dealing with the following three scenarios:
\begin{itemize}
	\item Pose transfer through extreme configurations: The planning and control motions remain feasible and efficient to large-range pose adjustment ($>100$ average distal joint position differences) near joint limits.
	\item Stable and reactive target contact: The robot motion is guided via a bidirectional manner, i.e. robot adaptively decides whether to reach or to stay clear from the target by evaluating the on-the-fly tool-target configuration.
	\item Smooth trajectory tracking: The robot is capable of following a predefined path with controller disturbances, where the tracking process automatically halts due to path deviation and will proceed after the deviation is cleared.
\end{itemize}

Basically, the framework provides a reactive planning and control process which could be used to characterized different types of elementary steps in surgical procedures, and could lead to a "fail-in-safety" performance which highly reacts to the on-the-fly information.
This is essential to further improve task-level reliability when automating a sequence of motion steps in order to complete a task successfully.
The quantified results will be later addressed in experiments.

\section{8. Experiments \& Results}

\begin{figure}[t]
\centering
\includegraphics[width=0.47\textwidth,keepaspectratio]{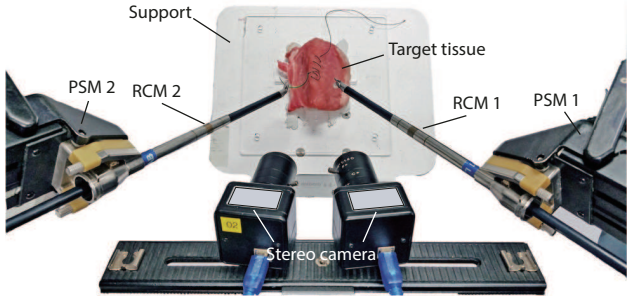}
\caption{The experimental set-up.}
\label{fig:ExpSetup}
\end{figure}

\begin{figure*}[t]
\centering
\includegraphics[width=0.92\textwidth,keepaspectratio]{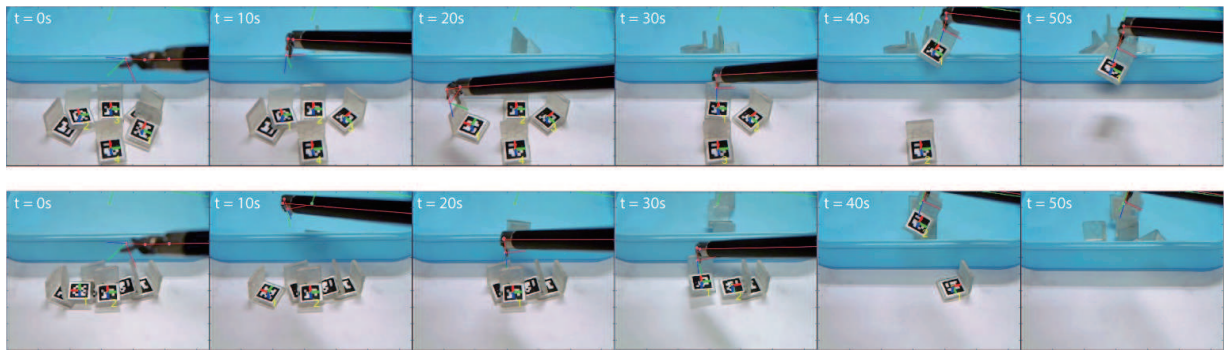}
\caption{Frames of the simulated debridement task using a single robotic instrument. The red and green robot skeleton represents the backprojected layout of the instrument's current and tissue releasing configuration on the camera image. The Cartesian frame of the instrument's end-effector (i.e. the tool) and the blocks are also visualized.}
\label{fig:ExpDebridementVideo}
\end{figure*}

\begin{figure}[t]
\centering
\includegraphics[width=0.4\textwidth,keepaspectratio]{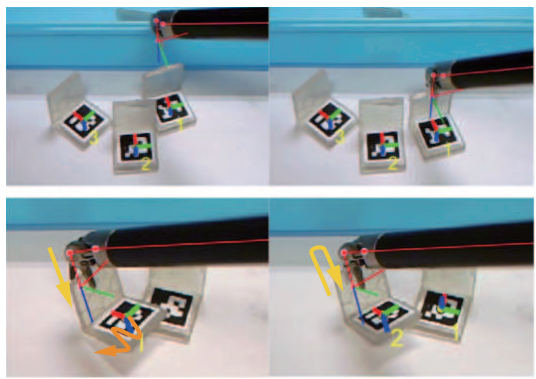}
\caption{Target uncertainty during task execution, where the block is unexpectedly deviated to a new position (deviated motion marked in orange arrows) due to premature tool-target collisions. The planned/replanned tool position is conceptually denoted as yellow arrows.}
\label{fig:ExpDebridementTest}
\end{figure}

\subsection{8.1. System Set-up}

In this section, we present results of on-site experiments for the robot to automatically perform simulated surgical tasks.
We using a dVRK with two PSMs as the robot platform.
Both PSMs are equipped with LNDs, which is a common type of instrument adopted when conducting tissue arrangement and suturing in RAMIS.
\hl{Both instruments have a 7-DoF joint set with the first four accounting for the RCM motions for minimally-invasive set-up, the middle two joints for providing wristed motions, and the last DoF for controlling the tool actuation (or the jaw opening angle).}
The physical setup is shown in Fig. \ref{fig:ExpSetup}.
We use a Core i7 3.4GHz user-side PC controller (with 156 GB RAM and without GPU acceleration) which connects to the robot controller with TCP/TP.
The cisst/SAW software environment is used and the algorithms run in MATLAB 2020b under the Robot Operating System (ROS).
A pair of industrial CMOS cameras are used for online visual feedback data acquisition with 640x480 resolution and 30 frames per second.
The base position of the two robots (i.e. the RCM) are set such that their distances to the target to be operated are both around 150 mm, which is commonly adopted in the clinical practice of RAMIS.
The camera is fixed between two PSMs, which is a typical set-up scale for multi-port RAMIS \shortcite{escobar2014atlas}.
The positions of the left camera relative to PSM1 and PSM2 are $[34.5, 70.2, 39.1]^{\intercal}$ mm and $[66.9, 104.6, 51.60]^{\intercal}$ mm, respectively.
The overall control loop of the system is around 20Hz without software acceleration.
The robot end-effector motions are saturated to 0.8 mm per step, which indicates a maximum 16 mm/s linear velocity for the tool.

\hl{We define the jaw angle to stay 0 during tool-centric actions because the $7^{th}$ is not considered in tool-centric actions.
While the instrument enters target-centric actions (with $\tau>0$ and $\dot{\tau}>0$), the jaw angle smoothly opens to a pre-defined angle ($30^{\circ}$) and will only close when $\tau\rightarrow1$.
When inverting the trajectory, the above sequence will be toggled accordingly.
This applies to any elementary motions which involve Mode II, III, IV.}

\hl{Several sensing algorithms are required for providing accurate 3D information to facilitate task completion is introduced in this subsection.
The focal lengths of the two cameras are tuned to $30$ m}m\footnote{The Karl Storz 1 S stereo laparoscope owns a focal length of $15-200$ mm}.
\hl{The image resolution of both cameras are set to 640$\times$480 which acquire video streams both at 30 frames per second.
The stereo camera has been calibrated in advance} \shortcite{zhang2000flexible} \hl{with its backprojection being 0.1582 pixel upon data acquisition of 30 pairs of sample images.}
\hl{The camera-robot transformation has been calibrated using the method based on our previous work} \shortcite{zhong2020hand} \hl{to all the engaged PSMs.
The performance of our framework is systematically
assessed by respectively performing single-arm tasks like debridement and membrane dissection, a dual-arm task like suturing.
Task effectiveness in terms of efficiency, success rate and accuracy will be validated.}

\subsection{8.3. Task I: Debridement}

The first scenario simulates the surgical tissue debridement procedure, which is designed to illustrate the performance of our framework to automate repetitive reach-and-grasp motions of the robotic instrument to different individual targets.
We fabricate a group of L-shape silicon gel blocks to simulate tissue debris.
Each block has a bounded 12 mm cube of size which simulates a plumped piece of tissue with a graspable edge, and is attached by a unique fiducial marker of AprilTag \shortcite{olson2011apriltag} with 25h9 tag family for 6-DoF target localization and multi-target identification.
The target is online detected and tracked during the task.
We apply the MoB to be Mode II (refer to Fig. \ref{fig:MoBTasks}) for automating the block reaching with grasping and Mode III for retracting with tissue releasing for debris collection to a plastic plate.

We design a multi-block debridement task to be completed by PSM 1.
The blocks are random placed within a rough volume of 50x50 mm$^{2}$ table surface, whose orientation are randomly settled as well.
As the blocks are identical, their poses are online tracked by the 
we calculate the constant grasping pose with respect to the fiducial marker prior to the start of the task.
Their pick-up sequence is then randomly assigned to avoid adding bias to statistical performance analysis.
The releasing pose $\textbf{T}_{rel}$ of the robot's end-effector for debris collection is assumed constant with 
\begin{equation} \label{eq:}
    \textbf{T}_{rel} = 
    \begin{bmatrix}
    	-0.924 & -0.222 & 0.310 & 0.117 \\
    	-0.176 & 0.970 & 0.170 & -0.050 \\ 
    	-0.338 & 0.103 & -0.936 & -0.0527 \\
    	0 & 0 & 0 & 1
    \end{bmatrix}
    ,
\end{equation}
i.e. 138.4 mm distance from the robot base.
The tool's opening DoF is controlled for after the current manipulation step is completed.

\begin{figure}[t]
\centering
\includegraphics[width=0.42\textwidth,keepaspectratio]{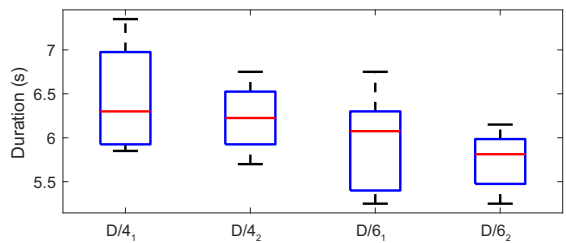}
\caption{Motion step durations during debridement, where the number 4$/$6 indicates the number of fragments, and the subscript 1$/$2 denotes the reaching$/$collecting motion of the fragment, respectively.}
\label{fig:ExpDebridementCatchBox}
\end{figure}

\begin{figure*}[t]
\centering
\includegraphics[width=0.92\textwidth,keepaspectratio]{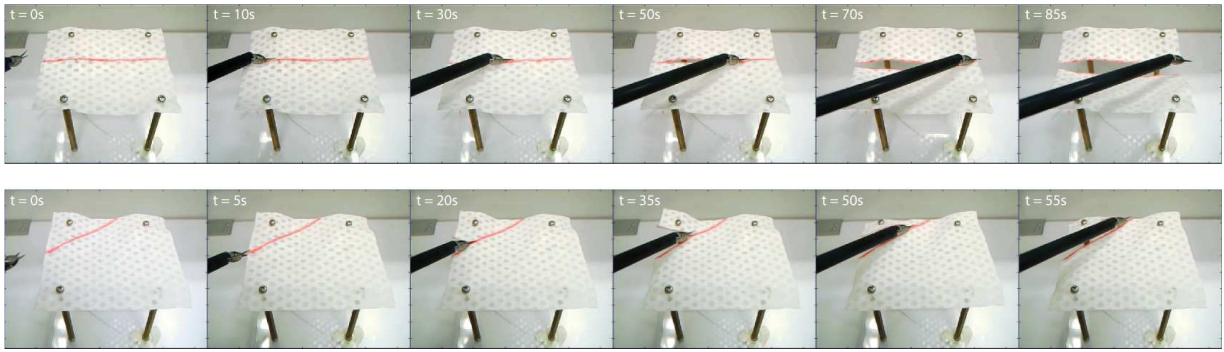}
\caption{Frames of automated phantom layer dissection procedures. The first and second row indicate the 90 mm case and 60 mm case, respectively, through paths with different directions.}
\label{fig:ExpCuttingVideo}
\end{figure*}

\begin{figure}[t]
\centering
\includegraphics[width=0.44\textwidth,keepaspectratio]{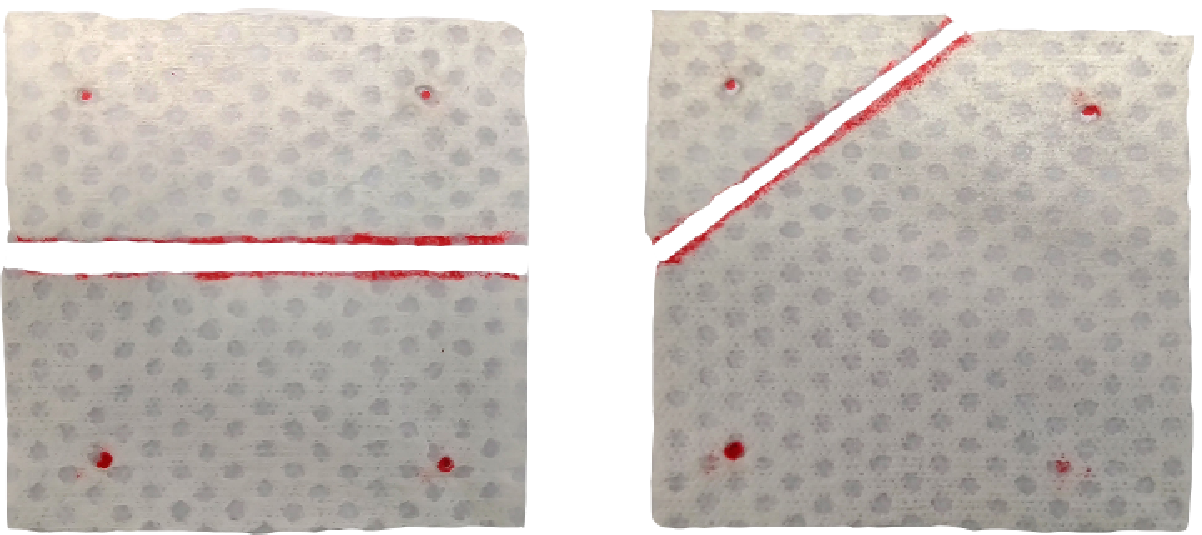}
\caption{Demonstration so the dissected phantom layers placed on the table with the prescribed cutting paths in red lines, and the fixation points to the support marked in red dots.}
\label{fig:ExpCuttingResult}
\end{figure}

We have conducted ten trials of six-block debridement and ten trials for four blocks.
As the main goal of the task is to clear the blocks from their initial positions, we evaluate the quality of this task by assessing the overall success rate of the whole task and the time distribution for each pick-up step.
Fig. \ref{fig:ExpDebridementVideo} illustrates the snapshots of two trials of six-block trials.
The overall success rate for the six-block and the four-block cases are all 90$\%$, i.e. both 9 out of 10.
Note that for each case, we rearrange the blocks by random placement.
Despite different poses, the instrument manages to collect all blocks smoothly.
The algorithm could guide the instrument adjacent to the block first and then finalize the contact reaching to achieve a safe grasp thanks to the GVM scheme.
It avoids premature tool-target collision during tool-centric action phase which can prevent the tissue from being dislocated or being damaged unexpectedly.
This is important as it is normally difficult to track the instant change of tissue's position shortly before contact due to visual occlusion.
We also show in Fig, \ref{fig:ExpDebridementTest} that during the grasping attempt, inaccurate instrument localization might unexpectedly deviate the target from its initial position.
Our framework could rapidly react by stating clear from the target first and initiates further adjustment.
As long as the target is detectable, the instrument will start another attempt once the tool-target configuration is desirable again.
Fig. \ref{fig:ExpDebridementCatchBox} shows the average execution time of each motion step during the task.
The median duration for collecting each block is 12.8/11.9 s for four/six block trials.
The four-block cases take longer time to complete one debris collection, as the more scattered placement of the blocks leads to longer instrument travel distances.
The random block setting does not significant affect the completion time for each attempt, which shows good consistency.

\begin{figure*}[t]
\centering
\includegraphics[width=0.94\textwidth,keepaspectratio]{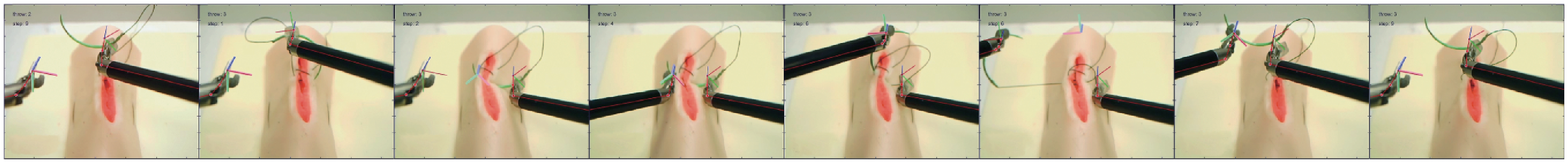}
\caption{Step illustration of a single-throw suturing during task automation. From left to right: 1) idle (the start of a throw); 2) needle (pre-entry) targeting; 3) needle insertion; 4) needle regrasping; 5) needle exiting; 6) suture pulling; 7) needle hand-off; 8) resetting (for starting the next throw).}
\label{fig:ExpSuturingStep}
\end{figure*}

\begin{figure}[t]
\centering
\includegraphics[width=0.42\textwidth,keepaspectratio]{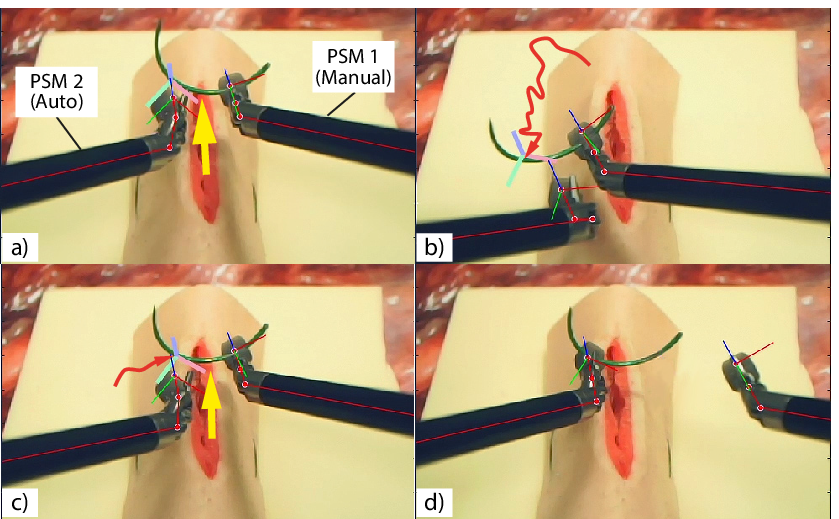}
\caption{When PSM1 is manually controlled by a user (with the shaky trajectory in red arrows), PSM2 stably tracks and makes contact attempt (in yellow arrow) when the tool-target alignment becomes desirable.}
\label{fig:ExpSharedVideo}
\end{figure}

\subsection{8.4. Task II: Tissue Membrane Dissection}

We next set up an experiment to simulate the dissection task to a tissue membrane, which is a typically demanded in surgeries involving organ removal like laparoscopic cholecystectomy \shortcite{reynolds2001first}.
The task includes highly repetitive cutting sub-steps in order to dissect the superficial layer of the tissue via a prescribed trajectory.
We use a soft handkerchief paper with average thickness of 1 mm as a phantom tissue layer.
The layer is fixed by a static support which keeps the tension of the surface to simulate the connectivity of the target to surrounding tissues and to reduce tissue deformation.
PSM2 is used to perform the task using the Potts Scissors instrument to provide cutting shear force to the target.
The tissue is of size 90 mm * 80 mm, and the dissection path on the tissue is marked by a red line that goes across the layer surface.
We set two types of path, one starts from the short edge and ends at the opposite one (with total length of 90 mm), and the other spans diagonally to the midpoint of the long edge (with totally length of around 60 mm).
The 3D positions of the path endpoints are detected first using the stereo camera and then are used to interpolate the path profile by comparing the backprojection error with respect to the observed one.
Finally, and the interval distance $d_{c}$ to between consecutive cutting steps is computed to determine the cutting points.
For empirical experiences, we select such distance to be $d_{c}=3$ mm, which indicates a 30-time cutting motion sequence for a 90 mm path and 20 cuts for 60 mm path.

We have conducted five consecutive trials for this task.
Three trials are set
We use Mode III to perform the dissection procedure (could be also solved by Mode IV but slightly slower) and the clamping motion of the scissors is done after settling the manipulation.
The setting of $d_{c}$ is set smaller than the length of the scissors (around 9 mm) allows a backward shifting before proceeding to the next step to avoid the tissue-tool adhesivity to after each cutting.
During the procedure, the instrument automatically plans and dissects the phantom layer progressively through repetitive cutting steps and separate the layer in half via the labelled paths.
Fig. \ref{fig:ExpCuttingVideo} shows the snapshots of the recorded clips of the real-time automated dissection process using a single instrument.
The robot successfully complete four out of five trials with the overall success rate thus being $80\%$.
The failure case is caused by tool-tissue entanglement with uncontrollable path deviation.
The average duration for executing the whole task is $81.1\pm1.2$ s and $54.3\pm0.9$ s for the 90 mm case and 50 mm case, respectively.
The time cost for performing each single cutting step is $3.9\pm0.3$ s.
The dissected layout of the phantom tissue layers are shown in Fig. \ref{fig:ExpCuttingResult}.
Upon manual measurement, the maximum lateral deviation from the prescribed linear path is around 1.5 mm for both 50 mm and 90 mm case (we assume the highest resolution for measurement is 0.5 mm as the smallest scale of the used ruler is 1 mm).

\subsection{8.5. Task III: Wound Suturing}

We finally conduct autonomous dual-arm multi-throw wound suturing as a comprehensive evaluation of our framework to perform complicated tasks.
The task appears in most surgical interventions, where a suturing needle is guided through wound edges for tissue approximation and wound closure.
The guideline is characterized by our framework in Fig. \ref{fig:MoBTasks} into seven individual steps for each throw.

\subsubsection{8.5.1. Scenario Setting \\}

To construct the set up, we use two PSMs which are both mounted with LNDs as the surgical tools.
To regulate the in-hand needle grasping position, we adopt the work in \shortcite{sen2016automating} that uses 3D-printed PVS needle
holder mounted along the jaws to enhance needle manipulation accuracy.
The stereo camera is placed between the two distal tools as a three-point invasion in typical laparoscopy \shortcite{horgan2001robots}.

Two types of tissues are used in our experiments.
The first type is the artificial soft tissue that consists of two layers, the outer layer to simulate tissue skin and the inner layer for the dermal structure.
Both layers are made from synthetic jel.
We also prepare a piece of porcine tissue by using two separate parts to create a lumped wound for ease of selecting needle insertion orientation.
The tissues are both fixed on a support to prevent unnecessary movements.
The positions of the wound edges are computed from the user-input wound endpoints via stereo images.
In addition, we introduce two parameters: the stitching width $d_{w}$ that denotes the distance between the needle's entry and exit point on the tissue, and the throw distance $d_{t}$ that denotes the (ideal) parallel distance of the suture between consecutive throws.
Without loss of generality, we set $d_{w}=10$ mm and $d_{t}=5$ mm as a resembling scale setting in standard robot-assisted surgical training procedures \shortcite{garcia1998manual}.

\begin{figure*}[t]
\centering
\includegraphics[width=0.94\textwidth,keepaspectratio]{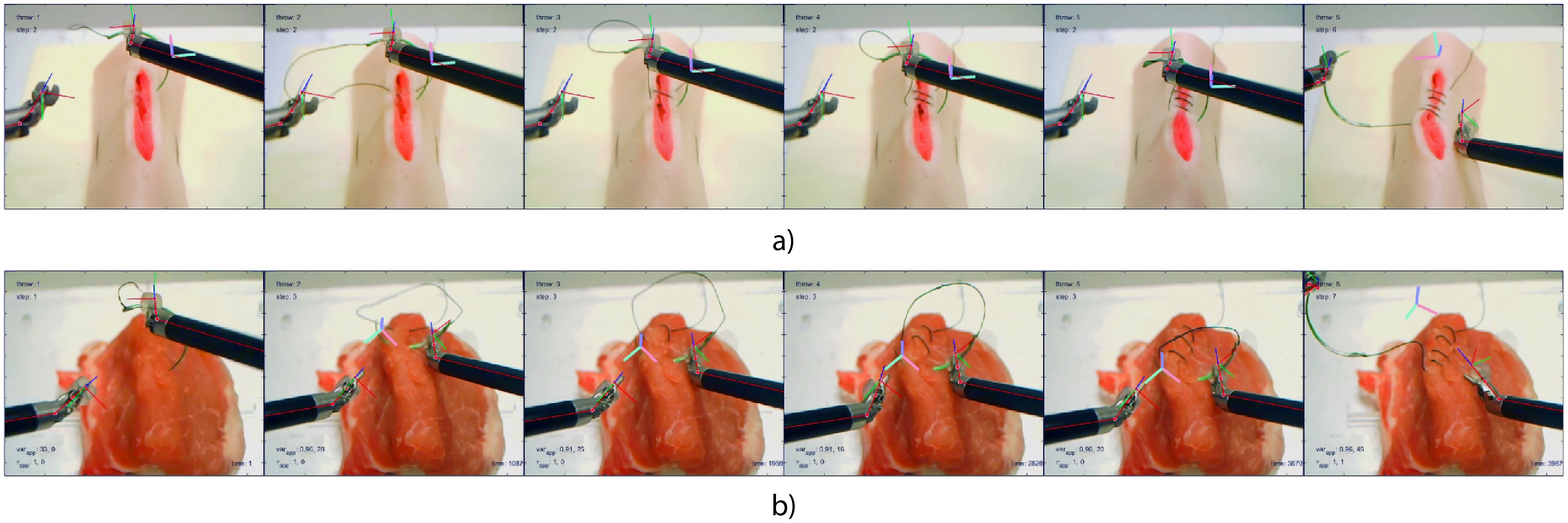}
\caption{Frames of dual-arm wound suturing using our framework on the artificial tissue (first row) and the porcine tissue (second row). Each image is captured at the same moment in the step (during needle targeting in the first row and during needle insertion in the next row).}
\label{fig:ExpSuturingVideo}
\end{figure*}

\begin{figure}[t]
\centering
\includegraphics[width=0.42\textwidth,keepaspectratio]{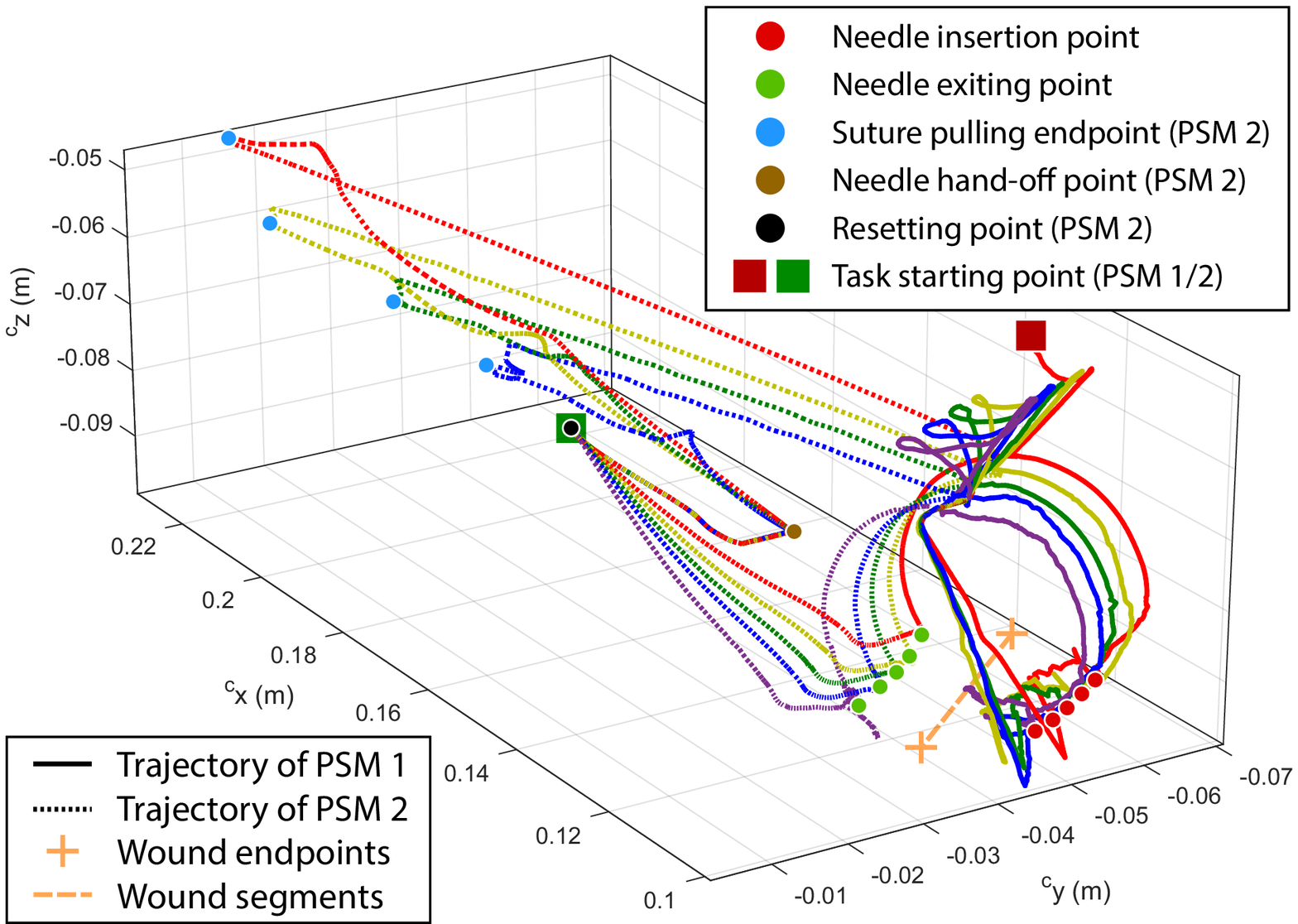}
\caption{Trajectories of the two instruments during  the five-throw suturing task, where the red/yellow/green/blue/purple indicates the trajectories from 1st to 5th throw, respectively.}
\label{fig:ExpSuturingTraj}
\end{figure}

\subsubsection{8.5.2. Single-throw suturing \\}

We first show the capability of our framework to complete a single-throw pipeline in suturing.
Fig. \ref{fig:ExpSuturingStep} shows the snapshots of individual motion steps of a single-throw pipeline on the artificial tissue.
The needle's insertion and exiting point on the tissue is directly assigned by user input that yields the preset $d_{w}$ and $d_{t}$.
The instruments start from preset idle configurations which are cleared from the target tissue.
The manipulation in each step is converged to the goal configuration with $\vert\vert\boldsymbol{\eta}\vert\vert$ down to a scale $\epsilon_{g}$ before proceeding to the next step.
To perform target-centric action to the needle during targeting, the tool-needle transformation is computed using image-based needle pose detection at the start of a throw.
During needle hand-off, two instruments work in a decentralized manner, where PSM2 guides the needle to a precomputed hand-off pose on top of the wound, with PSM1 automatically track the motion of PSM2 to decide a proper hand-off configuration subject to GVM scheme.
An example is particularly shown in Fig. \ref{fig:ExpSharedVideo} that demonstrates the decentralized needle hand-off process by human-robot collaboration.
The complete parameter settings for this scenario is shown in Table \ref{T3}.
Note that the pipeline in a single throw covers target contact (e.g. grasping the needle), trajectory tracking (e.g. needle insertion/existing), and dual-arm coordination (e.g. needle hand-off).
\hl{Due to the theoretical proofs of trajectory feasibility and motion stability, all the elementary motions are characterized and automated, involving target contact (including needle insertion, re-grasping, hand-off, etc.) while the path avoids premature collisions to the target tissue (refer to our supplementary video for demonstration).}

\begin{table*}[t]
\footnotesize\sf
\caption{Task-relevant parameters of our framework for multi-throw suturing. \label{T3}}
\begin{tabularx}{0.99\textwidth}{ccccc}
\toprule
~~~ & Pos. of idled PSM1$/$2 & Pos. of first insertion/exiting point & Pos. of last insertion$/$exiting point & Pos. of wound endpoints \\
\midrule
Artificial tissue
&
\makecell[c]{\text{[0.131   -0.067   -0.054]}$^{\intercal}$ \\ \text{[-0.091   -0.012   -0.045]}$^{\intercal}$}
&
\makecell[c]{\text{[0.117   -0.066   -0.096]}$^{\intercal}$ \\ \text{[-0.133   -0.044   -0.091]}$^{\intercal}$}
&
\makecell[c]{\text{[0.111   -0.053   -0.090]}$^{\intercal}$ \\ \text{[-0.122   -0.032   -0.095]}$^{\intercal}$}
&
\makecell[c]{\text{[0.130   -0.066   -0.093]}$^{\intercal}$ \\ \text{[0.116  -0.007   -0.060]}$^{\intercal}$} \\ \hline
Porcine tissue
&
\makecell[c]{\text{[0.116   -0.051   -0.055]}$^{\intercal}$ \\ \text{[-0.099   -0.044   -0.091]}$^{\intercal}$}
&
\makecell[c]{\text{[0.117   -0.074   -0.086]}$^{\intercal}$ \\ \text{[-0.140   -0.042   -0.079]}$^{\intercal}$}
& 
\makecell[c]{\text{[0.103   -0.053   -0.082]}$^{\intercal}$ \\ \text{[-0.123   -0.020   -0.078]}$^{\intercal}$}
&
\makecell[c]{\text{[0.131   -0.071   -0.083]}$^{\intercal}$ \\ \text{[0.119 -0.045 -0.081]}$^{\intercal}$} \\ \hline
\multicolumn{5}{l}{$^{*}$All positions (unit: m) are calculated relative to the respective robot base except the wound endpoints positions (from the camera).} \\ 
\bottomrule
\end{tabularx}
\end{table*}

\subsubsection{8.5.3. Multi-throw suturing \\}

We define a five-throw suturing task for artificial tissue and a four-throw suturing task on porcine tissue to comprehensively evaluate the task-level applicability of our framework.
The suturing will be performed using single-continuous suture pattern, which is a common and efficient type of suturing technique in RAMIS.
We do not include knot tying at the end of suturing due to the complexity of detecting the suture's 3D topology, which is not the main goal of this work.
The needle's insertion/exiting positions are computed based on two virtual 3D markers (named as the ``wound positioners'') manually input by the users via stereo images prior to the start of the task.
The needle insertion orientation is computable once the insertion and exiting point is known \shortcite{pedram2017autonomous}.
Note that we select $d_{t}$ as 8 mm for the porcine tissue to evaluate task performance under different settings.
The moving distance of the suture pulling step is computed based on its consumed length of suture in each throw, which is generically given as 14 mm measured from teleoperation based on the assigned $d_{w}$, $d_{t}$.
Trials on each type of tissue are conducted continuously to maintain identical setting throughout the experiments.

We perform ten trials for each type of tissue respectively under identical setting.
The task-level assessment include task success rate to evaluate reliability, duration of a single throw and the entire task for in-process performance, and resultant suture pattern accuracy for task quality.
Fig. \ref{fig:ExpSuturingVideo} illustrates the frames of task execution process on both the artificial and porcine tissue viewing from the sensing (left) camera.
The task is defined as ``successful'' only if the suture passes through both edges of the wound for enough times without user interference.
The overall success rate of performing five-throw suturing is $80\%$ on the artificial tissue (i.e. 8/10) and $70\%$ on the porcine tissue (i.e. 7/10).
The failure cases are either subject to needle-suture entanglement or missing the exiting point during insertion.
It could be further elevated to $90\%$ (i.e. 9/10) for both the artificial tissue and porcine tissue suturing, respectively, if manual suture arrangement is provided during the task.
Note that the above data is calculated by directly regarding a task as a failed trial once any step failure occurs during the process.
For step-wise failure, we manually halt the task once a failed step occurs and rearrange the needle/suture accordingly, and then let the robot proceed to the rest of the task to compute the accumulative number of step failure.
In this case, a total of 5 motion failures during 10 trials of five-throw suturing (or among the totally 640 motion steps), which indicates 0.8$\%$ of motion failure.

\begin{figure}[t]
\centering
\includegraphics[width=0.47\textwidth,keepaspectratio]{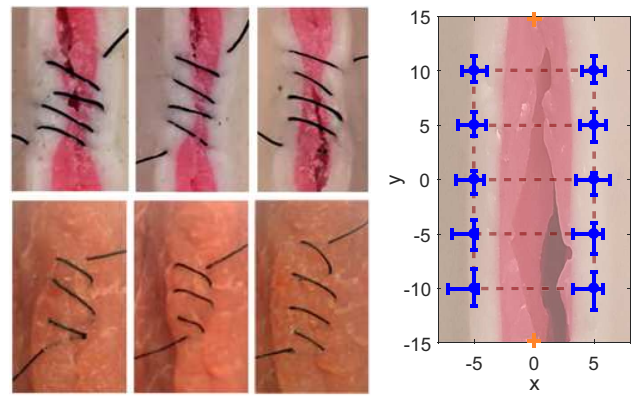}
\caption{The suture pattern on along the tissue wound edges after five-throw suturing (left-upper three: results on artificial tissue; left-bottom three: results on porcine tissue), and the manually measured suturing accuracy on the insertion/exiting positions (the right figure).}
\label{fig:ExpSuturingWound}
\end{figure}

\begin{figure}[t]
\centering
\includegraphics[width=0.45\textwidth,keepaspectratio]{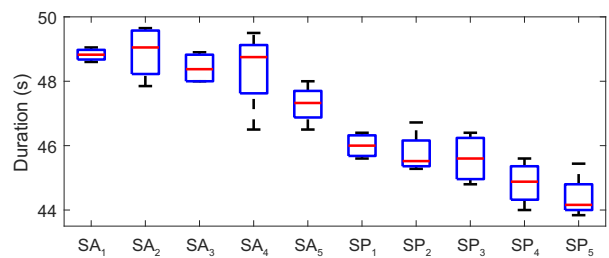}
\caption{Duration required for each throw for suturing an artificial tissue and a porcine tissue, respectively.
The gradual descent appears via consecutive throws in attributed to the shorter residual length of suture and thus shorter suture pulling distance under saturated tool velocity.}
\label{fig:ExpSuturingThrowBox}
\end{figure}

Suturing the porcine tissue owns lower success rate, as the tissue surface might encounter irreversible deformation after many stitches.
The trajectories of the two instruments' distal tool for completing such task is shown in Fig. \ref{fig:ExpSuturingTraj}.
It can be seen that the in-process trajectories maintain good consistency among individual throws without unpredictable movements.
Fig. \ref{fig:ExpSuturingWound} demonstrates the suturing accuracy on the artificial tissue\footnote{We only complete the suturing accuracy measurement to the trials targeting artificial tissue, as for the porcine tissue, it becomes difficult to find visual reference on its surface for measurement after the task.}.
Fig. \ref{fig:ExpSuturingThrowBox} shows the time cost for completing individual throws in the task.
The average time of performing one throw is 48.3$\pm$1.4 s and 45.1$\pm$1.0 s for artificial/porcine tissue, with the time variation percentage for being only 2.9$\%$ and 2.2$\%$, respectively, which shows good step-level consistency.
The average time cost for completing each throw gradually decreases (shown in Fig. \ref{fig:ExpSuturingThrowBox}), as the decreased residual suture length lowers the travel distance for suture pulling.
Furthermore, we characterize the outcome quality by manually measuring the deviation of the insertion/exiting position accuracy.
The total average accuracy is $1.7\pm1.1$ mm, with the maximum $2.0$ mm of wound deviation as the number of throws increases due to the irreversible deformation after several trials.
The deformation remains limited influence on suturing accuracy, and is regarded tolerable unless it causes misplacement of the needle during insertion (which is directly regarded as failure).

\subsection{8.6. Overall Assessment}

In this subsection, we validate the task-level performance of our framework by comparing with existing works reporting either manual task operations by novice surgeons, or state-of-the-art automation approaches.
\hl{We focus on aspects including the time cost, the step/task failure, number of individual trials in each task, etc. which are all metrics that reflect the task execution quality and consistency during performance assessment} \shortcite{martin1997objective}.
\hl{The existing works which aim to automate the identical tasks (e.g. continuous debridement, suturing) using the standard guidelines with articulated surgical robots will be especially selected and compared to ours as side-by-side analysis.
Note that we are unable to provide identical task-relevant parameter settings for performance comparison, as many works do not provide detailed parameter settings.
The comparison will be conducted based on their best reported results as well.}

\begin{table*}[h]
\centering
\footnotesize\sf
\caption{Comparison of manual/robot-assisted/autonomous tasks execution in physical environments. The time cost in blue font indicates the computed equivalent time cost for each elementary procedure in a task. Failure rate (column "Fail") is computed with average number of failures per task and/or total accumulative failure percentage (\textit{N.A.} indicates that the data is not given or uninterpretable). \label{T4}}
\begin{tabularx}{0.95\textwidth}{c|cclc|lllc}
\toprule
\multirow{3}{*}{Works} & \multicolumn{8}{c}{Tasks} \\
{} & \multicolumn{4}{c}{Debridement} \vline & \multicolumn{4}{c}{Suturing} \\
{} & Target & Qty. & ~Time cost (s) & Fail & ~~~~~Setting & Qty. & ~~~Time cost (s) & Fail \\ 
\midrule
\shortcite{garcia1998manual} & Bead & 10 & \makecell[l]{M~~~~~~~~68$\pm$32 \\~~~~~~~~~~~~~\blue{(6.8)} \\ RZ~~~~~183$\pm$42 \\~~~~~~~~~~~~\blue{(18.3)}} & \makecell[l]{5$\%$ \\ 3$\%$} & \makecell[l]{Needle~~~2-O \\ Width~~~~~10 mm} & 4 & \makecell[l]{M~~~~~~~~~154$\pm$40 \\~~~~~~~~~~~~~~\blue{(38.5)} \\ RZ~~~~~~~605$\pm$57 \\~~~~~~~~~~~~~\blue{(151.25)}} & \makecell[l]{8$\%$ \\ 8$\%$} \\ \hline

\shortcite{dakin2003comparison} & Peanut & 10 & \makecell[l]{M~~~~~~~~~~53.5 \\~~~~~~~~~~~~\blue{(5.35)} \\ RZ~~~~~~~128.5 \\~~~~~~~~~~~\blue{(12.85)} \\ RD~~~~~~~~61.0 \\~~~~~~~~~~~~~\blue{(6.1)}} & \makecell[l]{~~2.22 \\ (22.2$\%$) \\ ~~3.11 \\ (31.1$\%$) \\ ~~2.44 \\ (24.4$\%$)} & \makecell[l]{Needle~~~4-O \\ Width~~~~~5 mm} & 4 & \makecell[l]{M~~~~~~~~~~~~172.0 \\~~~~~~~~~~~~~~~\blue{(43.0)} \\ RZ~~~~~~~~~~426.1 \\~~~~~~~~~~~~~~\blue{(106.5)} \\ RD~~~~~~~~~~236.2 \\~~~~~~~~~~~~~~~\blue{(59.1)}} & \makecell[l]{~~3 \\ 2.56 \\ 1.78} \\ \hline

\shortcite{hubens2003performance} & Ring & 1 & \makecell[l]{M~~~~~~\blue{19.0$\pm$5.7} \\ RD~~~~~\blue{9.5$\pm$8.7}} & \makecell[l]{~2.25 \\ ~~~0 \\ {\tiny $*$median}} & \makecell[l]{Needle~~~4-O \\ Width~~~~~5 mm} & 4 & \makecell[l]{M~~~~~~356.4$\pm$112.6 \\~~~~~~~~~~~~~~~\blue{(89.1)} \\ RD~~~~~~60.5$\pm$39.2 \\~~~~~~~~~~~~~~~\blue{(15.2)}} & \makecell[l]{38 \\ ~2} \\ \hline

\shortcite{shah2009performance} & \multicolumn{4}{c}{\textit{N.A.}} \vline & \makecell[l]{Needle~~~20 mm \\ Width~~~~~5 mm} & 4 & \makecell[l]{RZ~~~~~~~~258$\pm$93 \\~~~~~~~~~~~~~~~\blue{(64.5)}} & \makecell[l]{~53} \\ \hline

\shortcite{fard2018automated} & \multicolumn{4}{c}{\textit{N.A.}} \vline & \makecell[l]{Needle~~~4-O \\ Width~~~~~5 mm} & 4 & \makecell[l]{RD~~~126.84$\pm$55.86 \\~~~~~~~~~~~~~~~\blue{(31.7)}} & \makecell[l]{\textit{N.A.}} \\ \hline

\shortcite{kehoe2014autonomous} & Foam & 1 & \makecell[l]{M~~~~~~~~~~\blue{29.0} \\ AR~~~~~~~~\blue{91.8}} & \makecell[l]{5.0$\%$ \\ 8.7$\%$} & \multicolumn{4}{c}{\textit{N.A.}} \\ \hline

\shortcite{mahler2014learning} & Foam & 1 & AR~~~~~~~~\blue{15.8} & \makecell[l]{\textit{N.A.}} & \multicolumn{4}{c}{\textit{N.A.}} \\ \hline

\shortcite{seita2018fast} & Seed & 8 & AD~~~~~~~\makecell[l]{57.62 \\ \blue{(7.20)}} & \makecell[l]{~~10 \\ (8.3$\%$)} & \multicolumn{4}{c}{\textit{N.A.}} \\ \hline

\shortcite{leonard2014smart} & \multicolumn{4}{c}{\textit{N.A.}} \vline & \makecell[l]{Needle~~~3-O \\ Width~~~~~10 mm} & 9 & \makecell[l]{M~~~~~~~560.4$\pm$358.6 \\~~~~~~~~~~~~~~~\blue{(62.3)} \\ RD~~~~~342.6$\pm$226.2 \\~~~~~~~~~~~~~~~~\blue{(38.1)} \\ A$*$~~~~~~~~64.51$\pm$0.81 \\~~~~~~~~~~~~~~~~~\blue{(7.2)}} & \textit{N.A.} \\ \hline

\shortcite{sen2016automating} & \multicolumn{4}{c}{\textit{N.A.}} \vline & \makecell[l]{Needle~~~39 mm \\ Width~~~~~5 mm} & 4 & \makecell[l]{M~~~~~~~~~~~~136.85 \\~~~~~~~~~~~~~~~~\blue{(34.2)} \\ AD~~~~~~~~~~383.00 \\~~~~~~~~~~~~~~~~\blue{(95.8)}} & 50$\%$ \\ \hline

\shortcite{pedram2020autonomous} & \multicolumn{4}{c}{\textit{N.A.}} \vline & \makecell[l]{Needle~~~30.55 mm \\ Width~~~~~16 mm} & 1 & AR~~~~~~~~~~~\blue{$\sim$200} & \makecell[l]{~\textit{N.A.}} \\ \hline

Ours & Syn-Jel & 6 & AD~~~~~\makecell[l]{48.8$\pm$1.3 \\ \blue{(6.2$\pm$0.3)}} & \makecell[l]{~0.1 \\ 1.7$\%$} & \makecell[l]{Needle~~~34 mm \\ Width~~~~~10 mm} & 5 & AD~~~~~~~~\makecell[l]{243.5$\pm$1.9 \\ \blue{(46.7$\pm$1.3)}} & ~\makecell[l]{~~0.5 \\ 1.4$\%$} \\ \hline
\multicolumn{8}{l}{Notation of task setting in the ``time cost'' columns.} \\
\multicolumn{8}{l}{First letter: 1) M: Manual execution with regular instruments; 2) R: Robot-assisted surgeon-centered execution; 3) A: Autonomy.} \\
\multicolumn{8}{l}{Second letter: 1) Z: Zeus robot; 2) D: dVSS$/$dVRK; 3) R: Raven Robot; 4) $*$: Other types of robots.} \\
\bottomrule
\end{tabularx}
\end{table*}

\begin{figure}[t]
\centering
\includegraphics[width=0.45\textwidth,keepaspectratio]{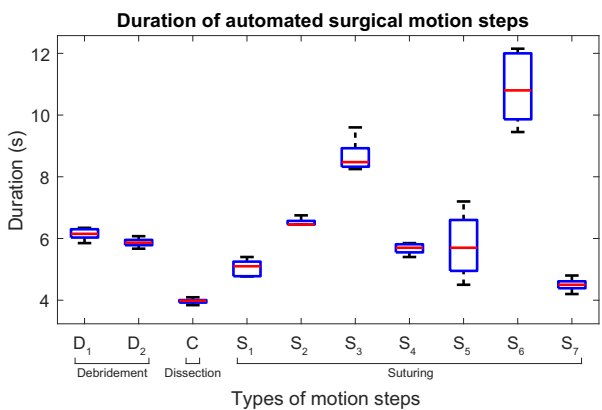}
\caption{Comparison of duration of individual motion steps in different surgical tasks (D indicates debridement task, C indicates dissection task, S indicates suturing task).}
\label{fig:ExpMotionStepBox}
\end{figure}

\begin{figure}[t]
\centering
\includegraphics[width=0.45\textwidth,keepaspectratio]{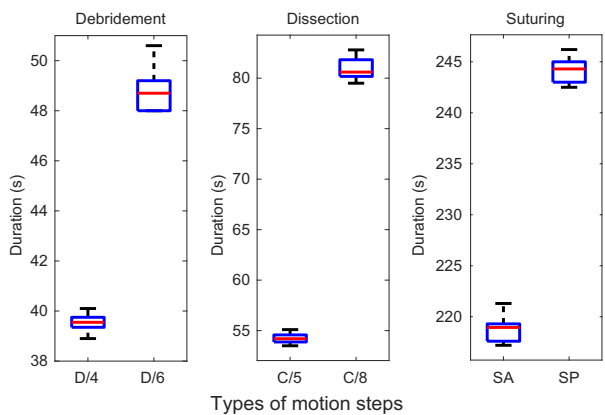}
\caption{\hl{Time consistency of completing task-relevant motions with different settings (D/4 and D/6 denote 4-/6-block debridement, C/5 and C/5 denote 50-/80-mm dissection path length, SA and SP denote suturing on artificial/porcine tissue, respectively.}}
\label{fig:ExpTaskOverallBox}
\end{figure}

We first summarize the duration performance from the above experiments into time cost for completing each task-relevant motion step (results shown in Fig. \ref{fig:ExpMotionStepBox}.
\hl{The time cost for completing the different task-relevant motions under different set-ups are shown in Fig. \ref{fig:ExpTaskOverallBox}.
We calculate the variation percentage of the duration for each step in terms of the average time cost is $3.3\%$/$4.5\%$ for the 4-/6-block debridement, $3.7\%$/$3.7\%$ for the 5/8 mm dissection procedure, and $1.8\%$/$1.4\%$ for the four-/five-throw suturing on the artificial/porcine tissue, respectively.
They indicate good consistency of robot's planning and control performance under different set-ups.
As there are current no existing works that automate different surgical tasks, the performance could be used as a benchmark for future works to refer to for multi-task automation.}

We have also listed results among different works in terms of set-ups, the time cost, and the accuracy (if available), where the results are shown in Table \ref{T4}.
For comparing task duration, we post-process the data of each work by computing the average time to execute one single step for each task (as shown in the data in blue font).
Our framework takes averagely 6.2$\pm$0.3 s to finish collecting a single fragment in debridement, which is slightly faster than many of the results using manual operations by novices, but a lot greater in time consistency.
Regarding suturing, our framework takes averagely 46.7$\pm$1.3 s for a single-throw suturing throw on the assigned tissue surface (maximum tool velocity currently set to 16mm/s), which is significantly more efficient (mostly 1.5-2 times faster) than manual operations, and the time variation is more than an order of magnitude smaller.
Note that we are also the first to conduct five-throw suturing on phantom tissue and also the first to use wristed robotic instruments for multi-throw suturing on porcine tissue.
The performance is mainly attributed to a close-form solution for integrated motion planning and control framework, which owns looping time of $<$1 ms without software acceleration, and doe not involve data-driven knowledge or iterative computations.
Although the works appeared in Table \ref{T4} own different set-ups and the performance might not be compared with identical scenarios, we show the capability of our work to be a baseline for generic surgery autonomy.

\section{9. Discussions and Conclusions}

\hl{In this paper, we have presented an generic planning and control framework to automate different types of surgical tasks.
The framework systematically addressed both surgery-specific motion safety and task-level characterization, which has been weakly explored.
The DS-based controller with SMA globally guarantees the step-level motion stability and trajectory reachability via Lyapunov stability using NSS as system states.}
\hl{The model is efficient, differentiable without iterations, and can solve constrained trajectories without high-dimensional workspace analysis.
Meanwhile, the GVM provides adaptive re-planning that guides the robot through constrained path to contact target properly based on on-the-fly situations.}
The framework was then extended to construct complete pipelines of various surgical tasks including single-instrument tasks like debridement, tissue dissection, and dual-instrument tasks like suturing.

The proposed framework has been validated through simulations for performance study, and then through experimental study on three different surgical tasks (but not limited to).
\hl{The performance comparison shows that our results own good task efficiency, reliability, and performance consistency that outperform manual operations by novice surgeons and state-of-the-art automation algorithms.
Despite the absence of detailed parameter settings provided by existing works, our still represents the farthest step of our
framework to generically handle continuous multi-step suturing and debridement.
The framework could be used when the camera is moving, which commonly appears in surgery, as long as the RCMs of the endoscopic camera and the instrument is not changed.
It could potentially be adopted to larger domain of robot-enabled surgical tasks that require more complicated end-effector motions with wristed profiles.}

We also address several limitations in this work.
We do not actively regulate the online tissue deformation induced by the task motions.
This is reported in our suturing experiments that might result in failure of the task due to the inaccurate needle insertion.
\hl{This could be solved by integrating tension information to formulate the tissue's dynamic change.
One could either directly measure the tool-tissue interaction force (requiring force sensors which could not be achieved by dVRK), or to adopt vision-based tracking of tissue deformation using trackable surface features (mostly individual points).
The deformation relative to its resting state could be online monitored to estimate the (relative) tension.
Any excessive deviations could be modelled as repellent factor to be avoided by the via pose in goal-varying manipulation.}
Meanwhile, the camera-robot transformation is pre-calculated using efficient autonomous calibration approach in \shortcite{zhong2020hand}.
\hl{It is not updated during the task, where the leading positioning error of the instrument (even though not significant) is not online refined, but currently being capable of obtaining good task success rate.
Once it could be updated to minimize the residual positioning error, the motion accuracy apart from the planning and control framework could be further improved, which is currently our ongoing work.}

There are several directions extendable from this work.
The task-level performance and assessment results could be used as a benchmark to be addressed by future approaches under such setting.
We are also developing learning-based perception algorithms to localize target surgical areas such that our framework could be applied to complex surgical environment that targets clinical applications.
Moreover, we would like to explore active tissue deformation regulation during task automation to further improve task-level reliability under dynamic/uncertain physical situations.

\begin{acks}
We acknowledge Mr. Dickson Chun Fung Li and Ms. Yaqing Wang for their assistance in the setup of physical experiments.

The Authors declare that there is no conflict of interest.
\end{acks}

\begin{funding}
This work is supported in part by the Shenzhen Portion of Shenzhen-Hong Kong Science and Technology Innovation Cooperation Zone under HZQB-KCZYB-20200089,  in part of the HK RGC under T42-409/18-R and 14202918,  in part by the Multi-Scale Medical Robotics Centre, InnoHK, and in part by the VC Fund 4930745 of the CUHK T Stone Robotics Institute.
\end{funding}

\section*{Appendix A. Anatomy of ``Weak Coupling'' on A Wristed Robot}

We take the EndoWrist robotic surgical instrument structure as an example.
Recall the robot kinematics in (\ref{eq:x}) and (\ref{eq:v}), we use the skeleton nodes in $\mathcal{N}=\{n_{1},n_{2},n_{3}\}$ to describe the 3D position of the end-effector as follows:
\begin{subequations} \label{eq:apx_1}
    \begin{align}
        \textbf{x}_{t} & = 
        \overrightarrow{\mathcal{O}n_{1}} +
        \overrightarrow{n_{1}n_{2}} +
        \overrightarrow{n_{2}n_{3}}  \label{eq:apx_1_1} \\
        & =
        \begin{bmatrix}
            q_{3} \\
            l_{w} \\
            l_{t} \\
        \end{bmatrix}
        ^{\intercal}
        \begin{bmatrix}
            \textbf{R}_{s}(\textbf{q}_{s})\overline{\textbf{r}}_{s} \\
            \textbf{R}_{w}(\textbf{q}_{s},\textbf{q}_{w})\overline{\textbf{r}}_{w} \\
            \textbf{R}_{t}(\textbf{q})\overline{\textbf{r}}_{t} \\
        \end{bmatrix}
        \label{eq:apx_1_2}
    \end{align}
\end{subequations}
where $l_{w/t}$ denote the link lengths of the instrument and $q_{3}$ is length of the shaft that passes the RCM,
$\overline{\textbf{r}}_{s/w/t}$ compute the directional vectors of the link's centerline fixed in local frame, $\textbf{x}_{t}$ is coincided with $\mathcal{O}_{e}$ due to standard DHc, $\textbf{R}(\cdot)$ denote the rotation matrices.
Differentiating (\ref{eq:apx_1}) yields:
%
\begin{equation} \label{eq:apx_2}
    \dot{\textbf{x}}_{t} =
    \begin{bmatrix}
         q_{3}\textbf{S}_{s}(\textbf{q}_{s}) + \textbf{R}_{s}(\textbf{q}_{s})\overline{\textbf{r}}_{s} \\
         l_{w}\textbf{R}_{s}{}^{s}\textbf{S}_{w}(\textbf{q}_{w})\overline{\textbf{r}}_{w} \\
         l_{t}\textbf{R}_{w}(\textbf{q}_{s},\textbf{q}_{w}){}^{w}\textbf{S}_{t}(q_{t})\overline{\textbf{r}}_{t}
    \end{bmatrix}
    ^{\intercal}
    \begin{bmatrix}
        \dot{\textbf{q}}_{s} \\
        \dot{\textbf{q}}_{w} \\
        \dot{q}_{t} \\
    \end{bmatrix} \\
\end{equation}
where $\textbf{S}(\cdot)$ are the local Jacobian matrices,
Rearrange (\ref{eq:apx_1_2}) further lead to the following factored form:
\begin{equation} \label{eq:apx_3}
    \dot{\textbf{x}}_{t} =
	\textbf{L}_{1}
	\textbf{A}_{1}(\textbf{q}_{s})
	\dot{\textbf{q}}_{s}
	+
	\textbf{L}_{2}
	\textbf{A}_{2}(\textbf{q}_{w})
	\dot{\textbf{q}}_{w}
	+
	\textbf{L}_{3}\textbf{A}_{3}(q_{t})
    \dot{q_{t}}
\end{equation}
where $\textbf{A}_{1/2/3}$ are the interaction matrices, with $\textbf{L}(\cdot)$ being
\begin{equation} \label{eq:apx_4}
	\mathbf{L}_{1} = 
		\begin{bmatrix}	
			l_{s} & 0 & 0 \\
			0 & l_{s} & 0 \\
			0 & 0 & 1 \\
		\end{bmatrix}
		,~~~
	\mathbf{L}_{2} = 
		\begin{bmatrix}	
			l_{w} & 0  \\
			0 & l_{w} 
		\end{bmatrix}
		,~~~
	\mathbf{L}_{3} = l_{t}
\end{equation}
which act as coefficient matrices during regulation of individual terms in (\ref{eq:apx_3}).
In robotic MIS, the recommended insertion length of the instrument shaft that passes the RCM (or the trocar entry), i.e. $q_{3}$, is around 150 mm to 200 mm \shortcite{escobar2014atlas}.
The links of the wristed structure (taking the Large Needle Driver provided by the dVSS) are $[l_{w},~l_{t}] = [9.1,~9.5]$ mm, respectively.
This indicates that practically, we can assume $q_{3}\gg l_{w,t}$, which will arise the following motion properties of the instrument upon such kinematic constraint:
\begin{itemize}
\setlength{\itemindent}{-1em}
    \item \textbf{Property 1}. The position of the end-effector $\textbf{x}_{t}(t)$ (i.e. $n_{1}$) is roughly estimated as $\textbf{x}_{s}(t)$ (the position of $n_{1}$) under any feasible $\textbf{q}(t)$:
    \begin{equation} \label{eq:apx_5}
        \textbf{x}_{t}(\textbf{q(t)})\approx\textbf{x}_{s}(\textbf{q}_{s})~\forall\textbf{q(t)}~~~s.t.~~q_{3}\gg l_{w,t}
    \end{equation}
    which could be easily derived from (\ref{eq:apx_1}) as the norm of $\overrightarrow{n_{1}n_{2}}$ and $\overrightarrow{n_{2}n_{3}}$ could be neglected.
    This also indicates that the adjustment of the end-effector position is dominated by $\textbf{q}_{s}$, as $\textbf{x}_{s}(\textbf{q}_{s})$.
    
    \item \textbf{Property 2}. If the end-effector motion space $\textbf{x}_{t}(\textbf{q})$ is highly restricted\footnote{This assumption is made by assuming that the lateral moving range of $\textbf{x}_{t}(\textbf{q})$ is much smaller than $\vert\vert\textbf{x}_{t}(\textbf{q})\vert\vert$, which is reasonable in the confined intra-corporeal surgical workspace, as it tolerates much smaller lateral end-effector motions compared to longitudinal motions.}, the rotation matrix $\textbf{R}_{s}(\textbf{q}_{s})$ could be considered unchanged, i.e.:
    \begin{equation} \label{eq:apx_6}
        \textbf{R}_{s}(\textbf{q}_{s})\approx\textbf{R}_{s}(\textbf{q}_{s_{0}})~\forall\textbf{q}~~~s.t.~~q_{3}\gg l_{w,t}
    \end{equation}
    as $\textbf{x}_{s}(\textbf{q}_{s})= \textbf{R}_{s}(\textbf{q}_{s})\overline{\textbf{r}}_{s}$ (\ref{eq:apx_5})) is also restricted to a small domain, leading $\dot{\textbf{R}}_{s}(\textbf{q}_{s})\rightarrow\textbf{I}_{3\times3}$.
    Here $\textbf{q}_{s_{0}}$ denotes the static initial configuration of the robot prior to the control.
    This also indicates that the adjustment of the end-effector rotation is dominated by $\textbf{q}_{w}$ and $q_{t}$, as $\textbf{R}_{t}(\textbf{q})\approx\textbf{R}_{s}(\textbf{q}_{s_{0}})\textbf{R}_{w}{}^{w}(\textbf{q}_{w})\textbf{R}_{t}(\textbf{q}_{t})$.
\end{itemize}
To summarize, the adjustment of the instrument's end-effector position and orientation are dominated by different (unique) set of robot joints in a relatively decoupled manner.
We name the them as the "weak coupling" effect during robot motion control and is naturally available to any serial robot manipulators with long proximal links and a wristed end-effector with short distal links (e.g. yielding $q_{3}\gg l_{w,t}$ in this case).
However, it should be noted that we only utilize these approximations to design a well-performed robot controller instead of direct analytical computation, as $q_{3}<100l_{w,t}$.

\bibliography{IJRR22}
\bibliographystyle{mslapa}

\end{document}